\documentclass[lettersize,journal]{IEEEtran}
\usepackage{amsmath,amsfonts}
\usepackage{algorithmic}
\usepackage{algorithm}
\usepackage{array}
\usepackage{subfigure}
\usepackage[caption=false,font=normalsize,labelfont=sf,textfont=sf]{subfig}
\usepackage{textcomp}
\usepackage{stfloats}
\usepackage{url}
\usepackage{verbatim}
\usepackage{graphicx}
\usepackage{cite}
\usepackage{multirow}
\usepackage[table]{xcolor}
\usepackage{amssymb}
\usepackage[inkscapelatex=false]{svg}
\usepackage{makecell}

\begin{document}

\title{Spatial Degradation-Aware and Temporal Consistent Diffusion Model for Compressed Video Super-Resolution}

\author{Hongyu An, Xinfeng Zhang*,~\IEEEmembership{Senior Member,~IEEE}, Shijie Zhao, Li Zhang,~\IEEEmembership{Senior Member,~IEEE},\\ and Ruiqin Xiong,~\IEEEmembership{Senior Member,~IEEE}
\thanks{* Corresponding author.}
\thanks{Hongyu An and Xinfeng Zhang are with the School of Computer Science and Technology, University of Chinese Academy of Sciences, Beijing 100190, China (e-mail: anhongyu22@mails.ucas.ac.cn; xfzhang@ucas.ac.cn).

Shijie Zhao is with ByteDance Inc., Shenzhen 518055, China (e-mail:
zhaoshijie.0526@bytedance.com).

Li Zhang is with ByteDance Inc., San Diego, CA 92121 USA (e-mail: lizhang.idm@bytedance.com).

Ruiqin Xiong is with the Institute of Digital Media, School of Electronic
Engineering and Computer Science, Peking University, Beijing 100871, China (e-mail: rqxiong@pku.edu.cn).
} 
}



\maketitle

\begin{abstract}
Due to storage and bandwidth limitations, videos transmitted over the Internet often exhibit low quality, characterized by low-resolution and compression artifacts. Although video super-resolution (VSR) is an efficient video enhancing technique, existing VSR methods focus less on compressed videos. Consequently, directly applying general VSR approaches fails to improve practical videos with compression artifacts, especially when frames are highly compressed at a low bit rate. The inevitable quantization information loss complicates the reconstruction of texture details. Recently, diffusion models have shown superior performance in low-level visual tasks. Leveraging the high-realism generation capability of diffusion models, we propose a novel method that exploits the priors of pre-trained diffusion models for compressed VSR. To mitigate spatial distortions and refine temporal consistency, we introduce a Spatial Degradation-Aware and Temporal Consistent (SDATC) diffusion model. Specifically, we incorporate a distortion control module (DCM) to modulate diffusion model inputs, thereby minimizing the impact of noise from low-quality frames on the generation stage. Subsequently, the diffusion model performs a denoising process to generate details, guided by a fine-tuned compression-aware prompt module (CAPM) and a spatio-temporal attention module (STAM). CAPM dynamically encodes compression-related information into prompts, enabling the sampling process to adapt to different degradation levels. Meanwhile, STAM extends the spatial attention mechanism into the spatio-temporal dimension, effectively capturing temporal correlations. Additionally, we utilize optical flow-based alignment during each denoising step to enhance the smoothness of output videos. Extensive experimental results on benchmark datasets demonstrate the effectiveness of our proposed modules in restoring compressed videos.
\end{abstract}


\section{Introduction}

Constrained by high memory and transmission costs, videos are usually down-sampled and compressed to meet practical requirements. Striking a balance between the texture detail quality and the storage space and bandwidth limitations remains a significant challenge for video applications. Video super-resolution (VSR) has emerged as a widely adopted technique to enhance video quality by reconstructing continuous high-resolution (HR) frames from corresponding low-resolution (LR) counterparts. With the advent of deep learning, both sliding windows network-based \cite{VSRnet,VESPCN,DUF,EDVR,MTUDM,MuCAN,TDAN,DMBN,VRT} and recurrent network-based \cite{FRVSR,RLSP,RBPN,BasicVSR,BasicVSR++,TCNet,PSRT,CTVSR,MIA-VSR} VSR approaches have achieved remarkable progress. However, these methods largely overlook the unique characteristics of real-world videos that are commonly stored and delivered on the Internet or mobile devices. Such videos are typically subjected to varying degrees of compression \cite{CAVSR}. As a result, existing VSR methods may mistakenly treat compression artifacts as genuine textures and inadvertently amplify them during restoration. Moreover, non-adaptive VSR models are unable to account for different compression intensities, leading to blurred outcomes.

\begin{figure}[!t]
\centering
\subfigure[The visual comparison of LQ frame (Down-sampled and compressed), general SR frame (LDMs), and improved SR frame (SDATC).]
{\includegraphics[width=\columnwidth]{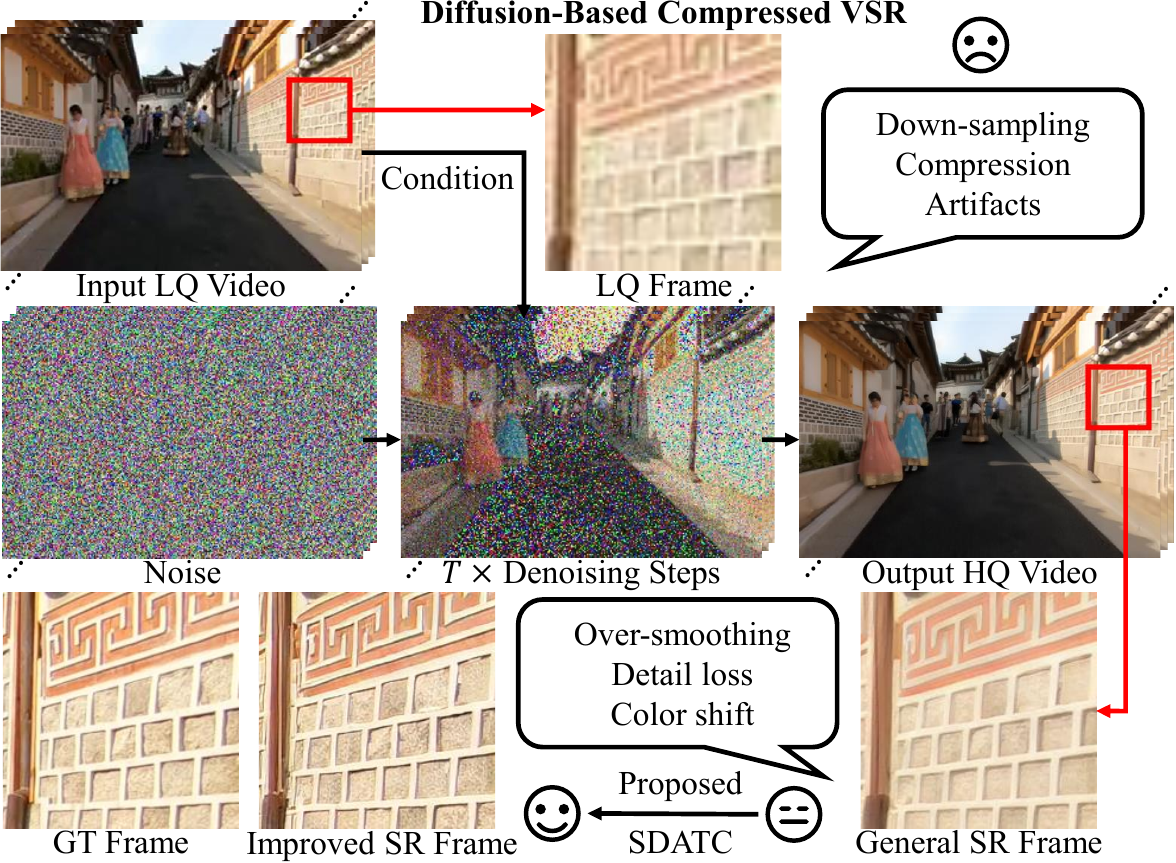}}
\subfigure[The LPIPS distance of diffusion models w/o and w/ the improvement.]{\includegraphics[width=\columnwidth]{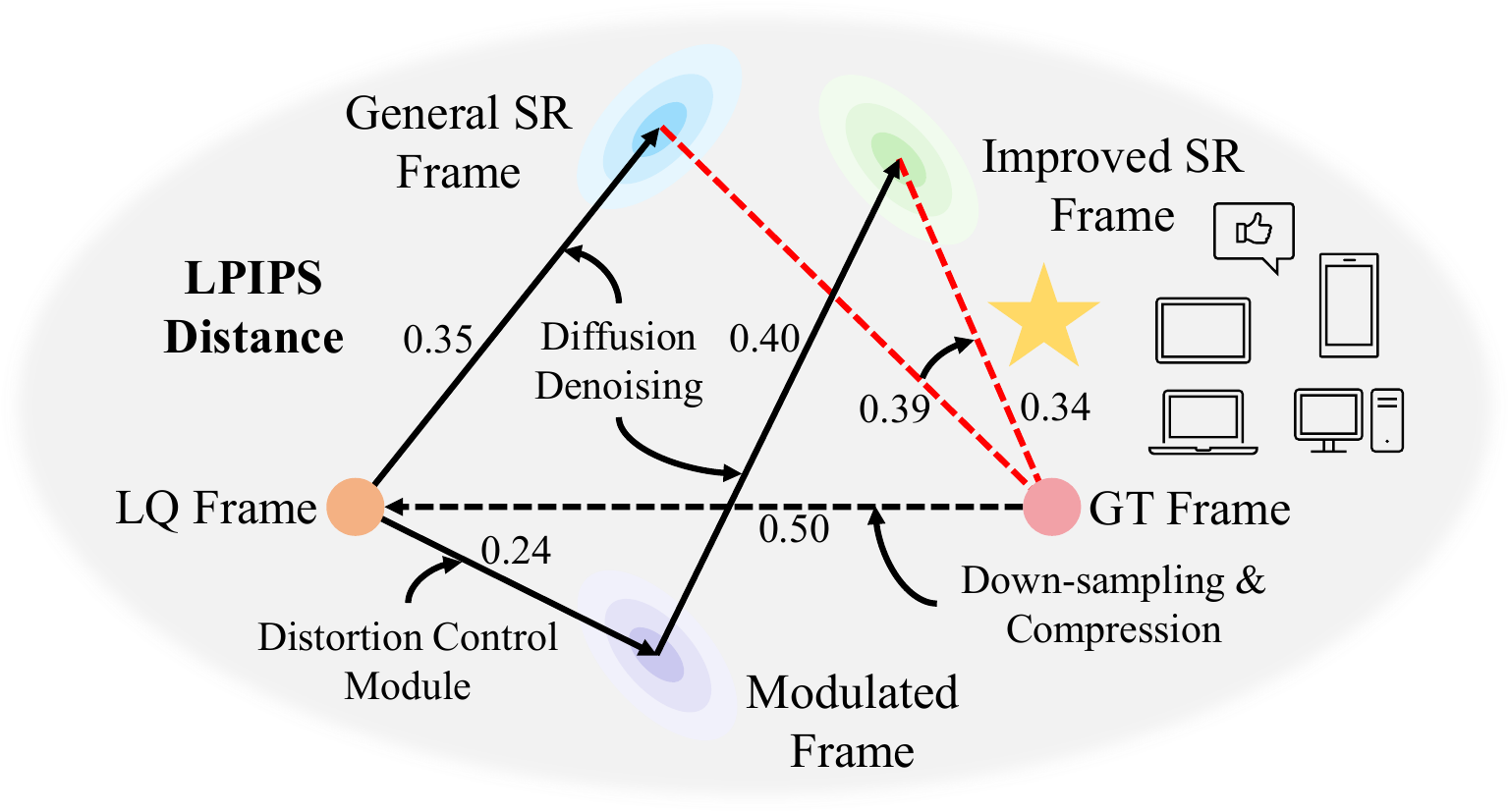}}
\vspace{-10pt}
\caption{The qualitative and quantitative comparison of our SDATC and other diffusion-based methods.}
\vspace{-10pt}
\label{fig.1}
\end{figure}

To enhance the resolution of degraded videos, several works have focused on compressed VSR. For instance, COMISR \cite{COMISR} exploited compression properties to mitigate distortions. FTVSR \cite{FTVSR} proposed a frequency Transformer architecture for compressed VSR. CAVSR \cite{CAVSR} utilized video stream information to predict compression ratios. Other methods modeled VSR with random distortions as a novel real-world super-resolution problem. Real-ESRGAN \cite{RealESRGAN} designed a high-order complex degradation simulation. RealBasicVSR \cite{RealBasicVSR} inserted a pre-cleaning stage to reduce artifacts during propagation.

While the aforementioned approaches improved compressed VSR performance through additional encoding priors or degradation simulations, restoring truncated texture details remains challenging. The quantization process during compression discards high-frequency information, leading to inevitable information loss. This lack of low-quality (LQ) video degradation priors hinders the reconstruction of visually pleasing results.

Inspired by the vivid generation capability of diffusion models, we exploit generative priors to address current challenges. Recent studies have successfully applied diffusion models to image super-resolution (SR), such as SR3 \cite{SR3} pioneered the usage of denoising diffusion probabilistic models (DDPM). StableSR \cite{StableSR} and DiffBIR \cite{DiffBIR} employed ControlNet \cite{ControlNet} to balance the realism and fidelity of recovered results. The following works \cite{CCSR,SeeSR,CoSeR,SUPIR,SSP-IR} tried to improve the generation process with multimodal prompts. Unfortunately, directly applying these image SR models with stochastic diffusion operations to compressed VSR can damage the temporal consistency and impair dynamic fluidity. Limited works \cite{StableVSR,SATeCo,Upscale-A-Video,MGLD-VSR} explored temporal alignment in diffusion-based VSR.

\begin{figure}[!t]
\centering
\subfigure[Image super-resolution diffusion model.]
{\includegraphics[width=\columnwidth]{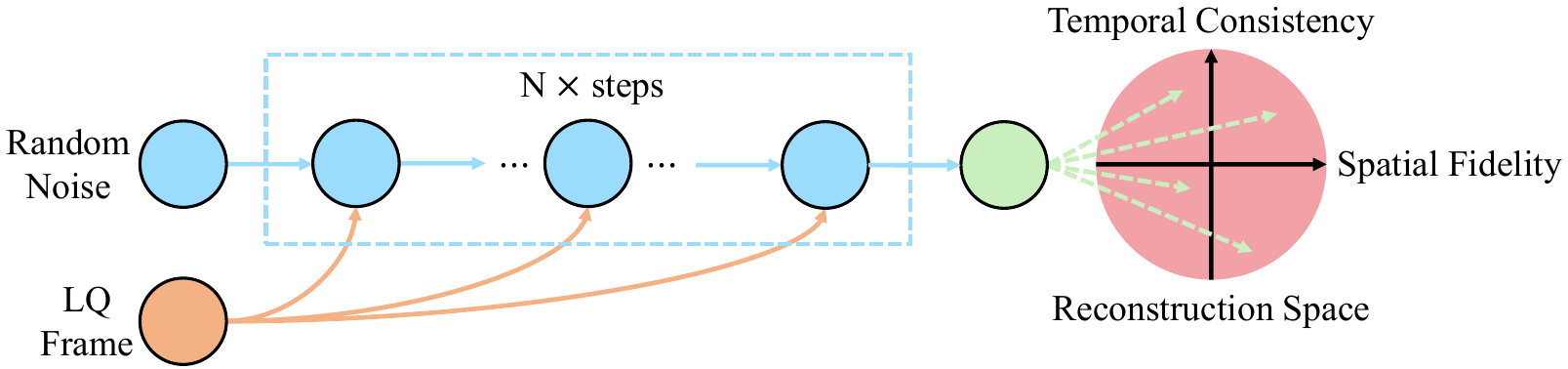}}
\subfigure[Spatial improved image super-resolution diffusion model.]{\includegraphics[width=\columnwidth]{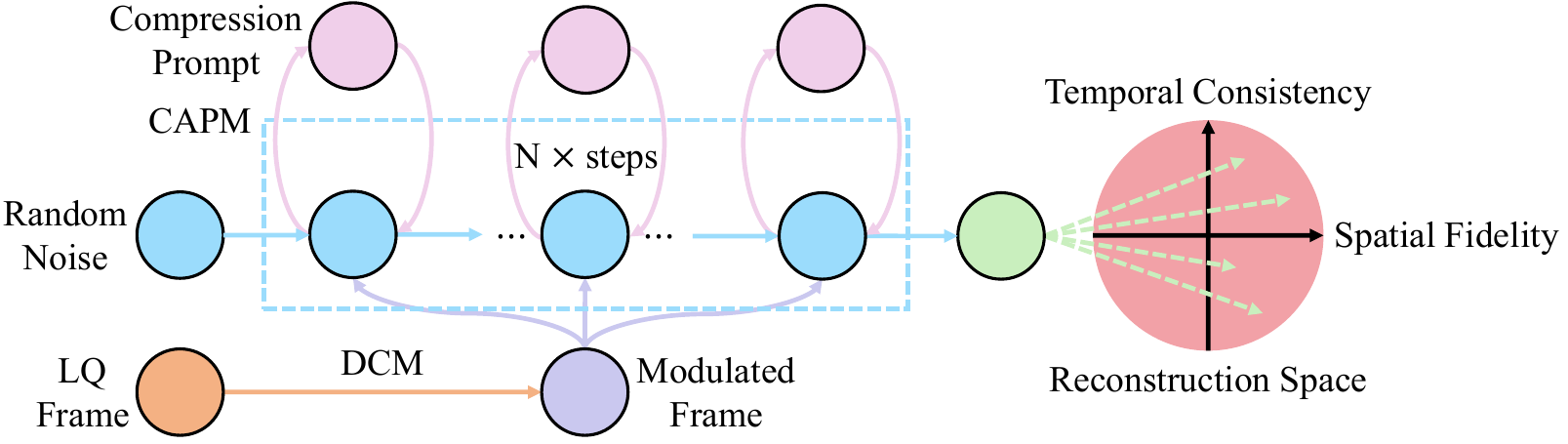}}
\subfigure[Temporal improved video super-resolution diffusion model.]{\includegraphics[width=\columnwidth]{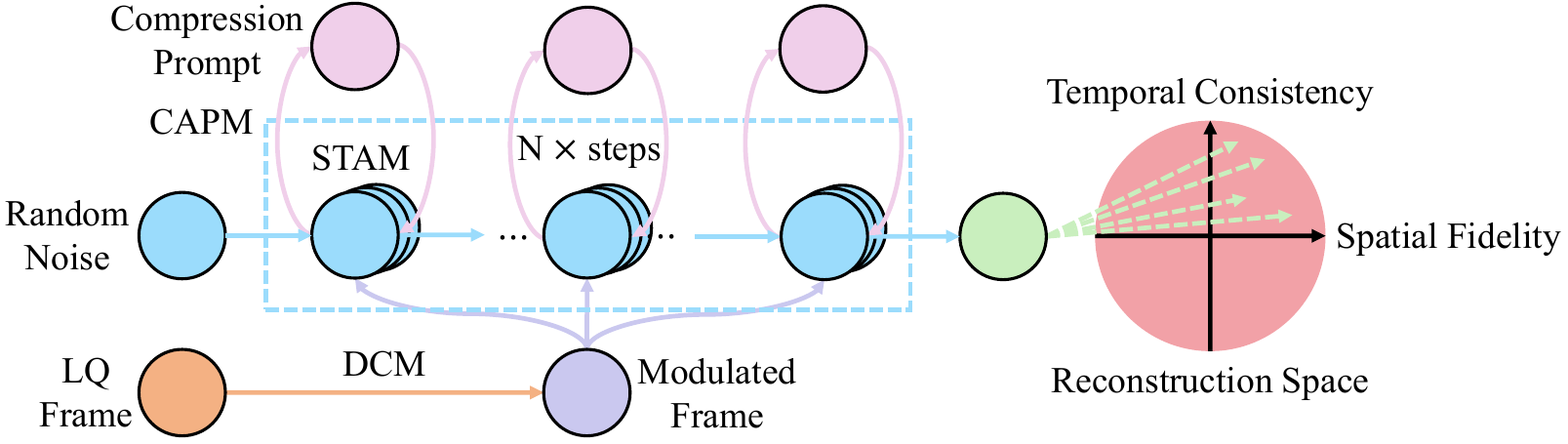}}
\vspace{-8pt}
\caption{Comparison of different diffusion processes for image/video super-resolution. Compared with (a) StableSR, our SDATC introduces spatial (b) and temporal (c) guidance for better generation.}
\vspace{-12pt}
\label{fig.2}
\end{figure}

The gaps between existing diffusion models and the compressed VSR task lie in two main aspects: (1) How to improve diffusion models to generate frames with higher spatial fidelity and fewer compression artifacts? (2) How to constrain the temporal consistency of reconstructed frames? To mitigate these gaps, we develop a distortion control module (DCM) to modulate the LQ inputs. The DCM eliminates interfering noise from the conditional LQ frames to enable more effective control of the following denoising phase. As illustrated in Fig. \ref{fig.1}(b), holding the diffusion model effectiveness constant, the designed pre-processing module prevents mistaken artifact generation and makes the SR distribution closer to the real domain, which improves the visual experience of output videos. Subsequently, we insert a compression-aware prompt module (CAPM) at UNet and VAE decoders to incorporate compression awareness. The UNet decoder accomplishes latent-space denoising and the VAE decoder completes pixel-space reconstruction. Based on the compression feature coding of different spatial distributions, CAPM provides lightweight prompts to characterize degradation information. Finally, we employ a spatio-temporal attention module (STAM) to explore relationships across frames with a spatial-temporal dimension fusion. The optical flow-based alignment during latent sampling also contributes to the continuous restoration.

In general, the proposed Spatial Degradation-Aware and Temporal Consistent (SDATC) diffusion model relieves negative compression impacts during spatial generation as shown in Fig.\ref{fig.2}(b). In contrast to current diffusion-based SR models, we divide the compressed VSR task into two sub-tasks. DCM preemptively eases degradation effects and PACM guides diffusion with compression-aware prompts. As depicted in Fig.\ref{fig.2}(c), STAM further takes full advantage of adjacent frames to smooth reconstructed frames.

The main technique contributions of this work can be summarized as follows:

\begin{itemize}
\item{We propose a distortion control module (DCM) to adjust the diffusion input and provide controllable guidance. The end-to-end DCM reduces content-independent degradations for the next generation procedure.}
\item{We introduce a compression-aware prompt module (CAPM) in UNet and VAE decoders to extract compression information from the latent and reconstruction space. CAPM enables an adaptive diffusion process for frames compressed to varying degrees.}
\item{We design a spatio-temporal attention module (STAM) and optical flow-based latent features warping to enhance temporal coherence.}
\item{Extensive experiments on various datasets with different compression levels demonstrate the superiority of our SDATC in terms of perception quality.}
\end{itemize}

\section{Related Work}

\subsection{Video Super-Resolution}
VSR exploits spatio-temporal similarity across LR videos to recover HR videos. VSR methods are commonly categorized into sliding-window \cite{VSRnet,VESPCN,DUF,EDVR,MTUDM,MuCAN,TDAN,DMBN,VRT} and recurrent frameworks \cite{FRVSR,RLSP,RBPN,BasicVSR,BasicVSR++,PSRT,TCNet,CTVSR,MIA-VSR}. The sliding-window framework processes reference frames within a moving window to recover target frames. VSRNet \cite{VSRnet} first employed a deep learning model for the VSR task. VESPCN \cite{VESPCN} introduced sub-pixel convolution for up-sampling frames and enhanced reconstruction through motion estimation. DUF \cite{DUF} applied 3D convolution to dynamically capture spatio-temporal relationships without motion compensation. DMBN \cite{DMBN} reduced the computational burden of 3D convolution. EDVR \cite{EDVR} proposed Deformable Convolutional Networks (DCN) \cite{DCN} based feature alignment and fusion. TDAN \cite{TDAN} further utilized DCN to estimate motion offsets between frames. MTUDM \cite{MTUDM} embed convolutional long-short-term memory to extract spatio-temporal correlations. VRT \cite{VRT} applied a Transformer-based recurrent framework.

The recurrent framework generates hidden states to convey long-range temporal information across frames. FRVSR \cite{FRVSR} integrated previous HR frames and subsequent LR frames to reconstruct target frames. RLSP \cite{RLSP} implicitly captures temporal relationships through hidden states. RBPN \cite{RBPN} concatenated outputs from a recurrent projection module to produce SR frames. BasicVSR \cite{BasicVSR} performed an optical flow-based bidirectional propagation mechanism to gather more information. The emerging BasicVSR++ \cite{BasicVSR++} carried out a bidirectional recurrent architecture and demonstrated promising performance. PSRT \cite{PSRT} analyzed alignment modules in Transformer-based architectures and proposed an efficient patch-level alignment. TCNet \cite{TCNet} further utilized a spatio-temporal stabilization module for frame alignment. CTVSR \cite{CTVSR} injected informative cues into a temporal trajectory to aggregate spatio-temporal correlations. MIA-VSR \cite{MIA-VSR} designed a masked intra-frame and inter-frame attention module to alleviate redundant computations by leveraging temporal continuity. However, the simple Bicubic down-sampling simulation used in these methods introduces synthetic-to-real gaps, which causes suboptimal performance in compressed VSR tasks.

\begin{figure*}[!t]
\centering
\includegraphics[width=1.95\columnwidth]{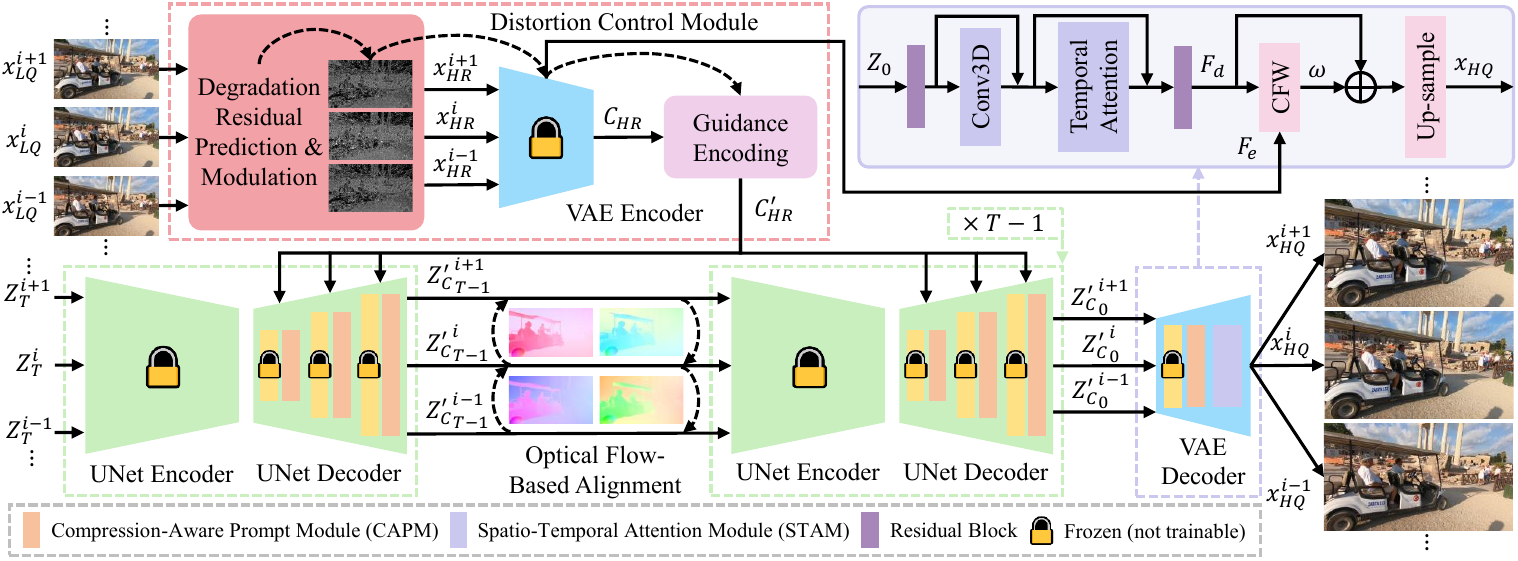}
\vspace{-6pt}
\caption{Overview of the proposed Spatial Degradation-Aware and Temporal Consistent (SDATC) diffusion model. We apply a distortion control module (DCM) to enhance input low-quality (LQ) frames. The modulated frames are fed into the Latent Diffusion Models (LDMs) based network as guidance. The trainable compression-aware prompt module (CAPM) catches degradation-specific details for generation. Moreover, we incorporate the fine-tuned spatio-temporal attention module (STAM) to preserve temporal consistency.}
\vspace{-10pt}
\label{fig.3}
\end{figure*}

\subsection{Compressed Video Super-Resolution}
The complex compression degradation poses new challenges for compressed VSR. To reduce artifacts, COMISR \cite{COMISR} enhanced the location and smoothness of compressed frames. FTVSR \cite{FTVSR} and its journal extension version \cite{FTVSR++} designed a DCT-based attention module to preserve high-frequency details. CAVSR \cite{CAVSR} estimated the compression level and applied corresponding treatments. Several works \cite{RealVSR,RealESRGAN,RealBasicVSR} tackled real-world VSR by synthesizing training data with mixed degradations. RealVSR \cite{RealVSR} collected paired LR-HR video sequences with the multi-camera system. Real-ESRGAN \cite{RealESRGAN} adopted a second-order process to flexibly mimic practical degradations. RealBasicVSR \cite{RealBasicVSR} proposed a stochastic degradation scheme and a real-world video benchmark dataset. The better compression estimation or more realistic degradation construction makes efforts on compressed VSR, but the information loss is difficult to recover with limited priors. When LR frames are compressed at low bit rates, the restored results are extremely blurry, making it difficult to discern and view.

\subsection{Diffusion-based Video Super-Resolution}
Diffusion-based image restoration has received increasing attention from researchers. The superior generative capability was explored in SISR \cite{SR3}. StableSR \cite{StableSR} and DiffBIR \cite{StableSR} used control modules during reconstruction. SeeSR \cite{SeeSR} presented a semantics-aware approach to preserve semantic fidelity in real-world image SR. SUPIR \cite{SUPIR} modified ControlNet \cite{ControlNet} and designed a novel ZeroSFT connector to reduce computational complexity, enabling a large-scale restoration model. SSP-IR \cite{SSP-IR} introduced an explicit-implicit strategy for semantic information extraction. In the domain of VSR, diffusion models have also shown promise. StableVSR \cite{StableVSR} exploited detail-rich and spatially-aligned texture information in adjacent frames. SATeCo \cite{SATeCo} pivoted on learning guidance from LR videos to calibrate spatio-temporal reconstruction. Upscale-A-Video \cite{Upscale-A-Video} introduced a flow-guided recurrent latent propagation module to enhance video stability. MGLD-VSR \cite{MGLD-VSR} proposed a diffusion sampling process based on motion-guided loss to generate temporally consistent latent features. In this work, we resolve the challenges of compressed VSR by leveraging the generative priors of pre-trained diffusion models. To overcome the limitations of existing diffusion frameworks, we develop spatial degradation-aware and temporal consistent techniques. These innovations could serve as a new paradigm for diffusion-based VSR models.

\section{Methodology}

Video compression standards, such as H.264 \cite{H.264}, trade off details for smaller file sizes, making it difficult to reconstruct realistic textures in compressed VSR. Motivated by the success of diffusion models, we exploit the generation priors in pre-trained Latent Diffusion Models (LDMs) \cite{LDM} to address this challenge. Unlike general diffusion models, LDMs apply Variational Auto-Encoder (VAE) to map images into latent space for decreasing training costs and enabling large-scale dataset application with rich prior knowledge. Nevertheless, the unstable LDMs generation can't handle compressed videos at unknown levels and increases temporal inconsistency.

To tackle these challenges, we propose a Spatial Degradation-Aware and Temporal Consistent (SDATC) diffusion model, which significantly restores clear videos and mitigates unpleasant artifacts. The overall framework of SDATC is demonstrated in Fig.\ref{fig.3}. Specifically, we fine-tune the UNet decoders in the down-sampled latent space and the VAE decoders in the pixel-level reconstruction space. Such a design with proposed modules effectively prevents a substantial increase in computational complexity while enhancing spatio-temporal SR performance. The architecture details of the proposed modules are presented in the following subsections. 

\subsection{Diffusion Model}
Inspired by principles from nonequilibrium thermodynamics, diffusion models generate images from random noise $z$ through an iterative reverse Markovian. The data distribution learning for generation is achieved through a $T$-step forward process. The diffusion from a clean image $x_0$ to Gaussian noise $x_T$ can be formulated as:
\begin{equation}
x_t=\sqrt{\alpha_t}x_{t-1} + \sqrt{1-\alpha_t}\epsilon_{t-1};x_t=\sqrt{\overline{\alpha}_t}x_{0} + \sqrt{1-\overline{\alpha}_t}\epsilon, 
\end{equation}
\begin{equation}
q(x_t|x_0)=N(x_t;\sqrt{\overline{\alpha}_t}x_{0},(1-\overline{\alpha}_t)I),
\end{equation}
where $t \in [1,T]$, $\epsilon \in N(0,1)$, and $\overline{\alpha}_t = \prod_{i=1}^{t} \alpha_i$. As $t$ increases, $\alpha_i$ gradually decreases, when $T \to \infty$, $x_T \in N(0,1)$. The reverse process predicts the inverse distribution based on the UNet network with the sampling process:
\begin{equation}
q(x_{t-1}|x_t,x_0)=N(x_{t-1};\tilde{\mu}(x_t,x_0),\tilde{\beta}_tI).
\end{equation}
The training goal is to obtain a denoising network $\epsilon_{\theta}$ by minimizing $\mathbb{E}_{t \in [1,T],x_0,\epsilon_t} [{\|\epsilon_t - \epsilon_{\theta}(x_t, t)\|}^2]$ to estimate the noise $\epsilon_t$. Based on the trained denoising network $\epsilon_{\theta}$, the model performs $T$ iterations of diffusion reverse denoising.

Building on the theoretical foundations, LDMs further utilize a pre-trained VQ-VAE \cite{VQ-VAE} to map the input image $x_0$ into a high-dimensional perceptual space. Given an input image $x_0 \in \mathbb{R}^{H \times W \times 3}$, the VQ-VAE compresses $x_0$ into a latent variable $\hat{z} \in \mathbb{R}^{h \times w \times d}$. The $h=H/4$ and $w=W/4$ are the scaled height and width, respectively, and $d$ is the refined dimension. The $\hat{z}$ undergoes $T$ diffusion steps and is subsequently decoded by the VQ-VAE decoder to produce the reconstructed frame $\hat{x}$. The proposed SDATC applies the pretrained Stable Diffusion v2.1 as its LDMs backbone. 

\subsection{Distortion Control Module}

Given a $n$ frames low-quality (LQ) video sequence ${\{x_{LQ}^1, ..., x_{LQ}^i, ..., x_{LQ}^n\}}$, we aim to recover a high quality (HQ) video sequence ${\{x_{HQ}^1, ..., x_{HQ}^i, ..., x_{HQ}^n\}}$. Most existing diffusion-based VSR methods first up-sample the input frames to the target resolution and then generate details guided by the up-sampled frames. However, commonly utilized up-sampling methods Bilinear and Bicubic can not estimate or modulate degradations. Therefore, the complex down-sampling and compression distortions in the input frames negatively impact subsequent generations by introducing mistaken priors.

\begin{figure}[!t]
\centering
\includegraphics[width=0.88\columnwidth]{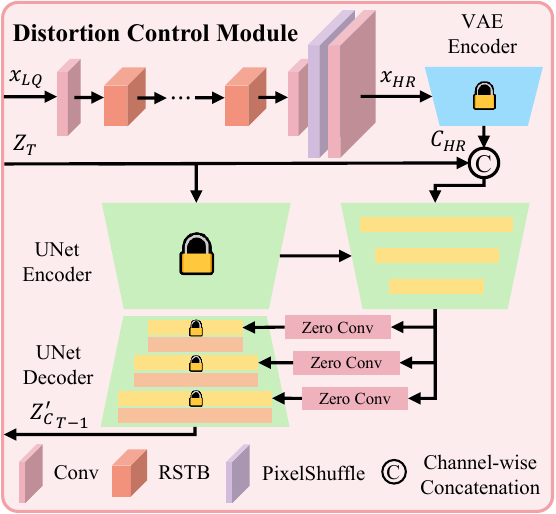}
\vspace{-6pt}
\caption{The distortion control module (DCM) designed as the pre-processing of the diffusion model framework.}
\vspace{-10pt}
\label{fig.4}
\end{figure}

To tackle the aforementioned issues and concentrate on compression characteristics, we design a distortion control module (DCM). Notably, LDMs learn the distribution of input data and original distortions in frames can interfere with the generation procedure, introducing unpleasant artifacts. To prevent noise corruption in LDMs, we employ the DCM to extract LQ guidance for the subsequent diffusion process. Specifically, we apply a Transformer-based network to remove distortions and increase spatial resolution as follows:
\begin{equation}
x_{\mathit{HR}}=\text{Up}(\text{RSTB}(\text{Conv}_{3\times3}(x_{\mathit{LQ}}))),
\end{equation}
where RSTB($\cdot$) depicts Residual Swin Transformer Blocks \cite{SwinIR} and Up($\cdot$) represents PixelShuffle up-sampling. Next, we encode the modulated $x_{\mathit{HR}}$ into the conditional latent space $\mathit{C_{HR}}$ through the VAE encoder. Following ControlNet \cite{ControlNet}, we concatenate the conditional guidance $\mathit{C_{HR}}$ with noise $Z_{t}$ and input it into a trainable copy of UNet encoder to obtain the guidance $C'_{HR}$. The fine-tuning of the UNet decoder with this guidance is then denoted as:
\begin{equation}
Z'_t=\text{UNet}_\text{decoder}(\text{Cat}(Z_t,\text{Conv}_\text{zero}(C'_{HR}))),
\end{equation}
To prevent early-stage random noise fluctuation, we introduce zero convolution. The proposed DCM is presented in Fig. \ref{fig.4}, we design a general pre-processing module for diffusion-based VSR and encode modified conditions to guide subsequent generation. Moreover, we fine-tune the DCM with LDMs in an end-to-end framework to optimize the input distribution.

\subsection{Compression-Aware Prompt Module}

Prompt learning has achieved great success in natural language processing by leveraging effective context information. Recently, PromptIR \cite{PromptIR} extended prompt learning to image restoration. Nevertheless, the prompts in PromptIR are randomly initialized and simply acquired through feature averaging and linear mapping, which limits the accuracy of degradation estimation. Consequently, we introduce an auxiliary encoder to extract compression representations. We predict accurate weights and assign prompts by transforming features into the degradation space. The proposed compression-aware prompt module (CAPM) is integrated and fine-tuned with both the UNet and VAE decoders to guide different representation spaces. Compared with the complex pre-training required for large-scale degradation datasets or the intricate semantic descriptions from large language models, CAPM is more feasible and computationally efficient, while effectively handling various compression levels.

\begin{figure}[!t]
\centering
\includegraphics[width=0.9\columnwidth]{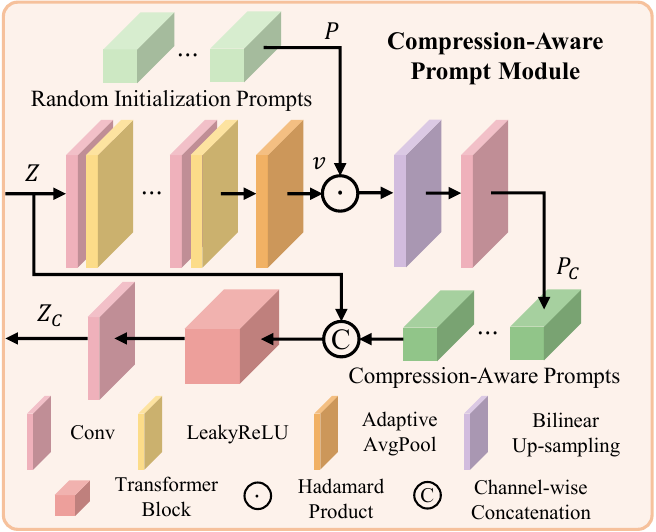}
\vspace{-6pt}
\caption{The compression-aware prompt module (CAPM). CAPM extracts compression information as prompts to direct better generation.}
\vspace{-14pt}
\label{fig.5}
\end{figure}

As depicted in Fig. \ref{fig.5}, CAPM transforms the latent feature $Z'$ into the compression space via an auxiliary encoder. The encoder comprises a CNN and an Adaptive AvgPool (AAP) layer, which encodes contextual compression priors in the feature vector $v$. We then apply $v$ to weight the prompt components $P$ and upscale $P$ to $P_{C}$, matching the size of the specific $Z'$. CAPM is inserted at each level of the UNet and VAE decoders to capture multi-scale correlations. The compression-aware prompts generation is summarized as:
\begin{equation}
P_{C}=\text{Conv}_{3\times3}\left(\sum_{k=1}^{K} {\text{AAP}(\text{Conv}_{3\times3}(Z))_kP_k}\right),
\end{equation}where $k$ denotes the length of the prompts. $P_{C}$ facilitates interaction between the latent feature $Z'$ and the prompt $P$ to extract compression information. Finally, we concatenate $P_{C}$ with $Z'$ and process compression-aware prompts through a Transformer block. The feature transformation can be formulated as follows:
\begin{equation}
Z'_{C}=\text{Conv}_{3\times3}(\text{Transformer}(\text{Cat}(Z, P_{C}))).
\end{equation} 

\subsection{Spatio-Temporal Attention Module}
Although LDMs-based SR methods can successfully reconstruct individual frames, multi-frame generation suffers from temporal inconsistency. Severe deformation of objects across adjacent frames is visually disruptive, and compression artifacts further exacerbate flickering. To improve compressed VSR with temporal consistency, we introduce a spatio-temporal attention module (STAM) in the VAE decoder. In particular, we expand the temporal dimension by incorporating multiple frames. Freezing the pre-trained spatial residual blocks, we insert 3D CNNs and temporal attention blocks (TAB). The TAB performs self-attention among the temporal dimension, and its outputs are regarded as residuals. We apply learnable parameters to balance spatio-temporal branches as:
\begin{equation}
Z_0'=\text{Res}(\text{Conv}_{3\times3}({Z'_{C}}_0)),
\end{equation}
\begin{equation}
Z_0''=\alpha_T\text{Conv3D}_{3\times3}(Z_0')+(1-\alpha_T)(Z_0'),
\end{equation}
\begin{equation}
F_d=\text{Res}(\beta_T\text{TA}(Z_0'')+(1-\beta_T)(Z_0'')),
\end{equation}where Res($\cdot$) is the residual block, $\alpha_T$ and $\beta_T$ denote learnable spatio-temporal tensors. As illustrated in Fig. \ref{fig.2}, we further incorporate the feature $F_e$ from the VAE encoder and achieve a balanced outcome through the Controllable Feature Warping (CFW) module \cite{StableSR}. The adjustable parameter $\omega$ (where a larger $\omega$ means higher fidelity) controls the reconstructed outputs as follows:
\begin{equation}
x_{HQ}=\text{Up}(F_d + \omega\text{CFW}(F_e, F_d)).
\end{equation} 

Due to computational constraints, we focus on enhancing temporal coherence within the VAE decoder. In the latent space of the UNet decoder, we compute forward and backward optical flow using RAFT \cite{RAFT} to align features and improve temporal consistency. As shown in Fig. \ref{fig.2}, we calculate the motion error $E_t$ at each denoising step:
\begin{equation}
E_t=\sum_{i=1}^{N-1} {\|f_{b}({Z'}_t^i)-{Z'}_t^{i+1}\|_1}+\sum_{i=2}^{N} {\|f_{f}({Z'}_t^{i})-{Z'}_t^{i-1}\|_1},
\end{equation}where $f_b$ and $f_f$ indicate backward and forward warping, respectively. The subsequent sampling process is as follows:
\begin{equation}
{Z'}_t=\text{UNet}({Z'}_{t+1})-\sigma_t^2 \triangledown _Z(\text{UNet}(Z^{'}_{t+1}), E_t).
\end{equation}The first item is the DDPM result, while the second term is the optical flow warping gradient scaled by variance $\sigma_t^2$. The gradient update in $\text{UNet}({Z'}_{t+1})$ is based on $E_t$.

\subsection{Color Correction}
Recent works \cite{color, StableSR} have identified that diffusion models encounter the issue of color shift. Notably, the variant network of diffusion models tends to exhibit a more pronounced color shift after training. To address this, Upscale-A-Video \cite{Upscale-A-Video} employs a wavelet color correction module \cite{StableSR} for correction. Specifically, Upscale-A-Video performs color normalization on the generated images by referencing the mean and variance of the LR inputs. Following a similar approach, we adopt adaptive instance normalization (adaIN) \cite{adaIN} to transform the style of the reconstructed frames, ensuring they have similar colors and illuminations to LQ frames.

\begin{table*}[!t]
\caption{Quantitative comparison of $\times$4 VSR on different compression level datasets. \colorbox{red!30}{\textbf{Bold}} and \colorbox{pink!50}{\underline{underlined}} values denote the best and second-best results respectively. $\uparrow$ and $\downarrow$ indicate better quality with higher and lower values correspondingly.}
\centering
\footnotesize
\vspace{-2pt}
\renewcommand\arraystretch{0.95}{
\setlength{\tabcolsep}{4.5pt}
\begin{tabular}{c|l|c|c|c|c|c|c|c|c|c}
\hline
\multicolumn{3}{c|}{Dataset}&\multicolumn{8}{c}{REDS4}\\
\hline
\multicolumn{2}{c|}{Metrics}&CRF&PSNR$\uparrow$&SSIM$\uparrow$&LPIPS$\downarrow$&DISTS$\downarrow$&FID$\downarrow$&NIQE$\downarrow$&MANIQA$\uparrow$&CLIP-IQA$\uparrow$\\
\hline
\multirow{5}{*}{\makecell{Non-generative \\ Methods}}&(CVPR'22) BasicVSR++\cite{BasicVSR++}&15&29.83&0.8184&0.3517&0.1148&38.57&5.017&0.2338&0.4984\\
&(CVPR'22) RealBasicVSR\cite{RealBasicVSR}&15&27.87&0.7786&\makecell[tc]{\cellcolor{red!30}\textbf{0.2689}}&\makecell[tc]{\cellcolor{pink!50}\underline{0.0684}}&36.08&2.647&0.3401&0.5295\\
&(ECCV'22) FTVSR\cite{FTVSR}&15&\makecell[tc]{\cellcolor{pink!50}\underline{30.88}}&\makecell[tc]{\cellcolor{pink!50}\underline{0.8580}}&0.3078&0.0991&33.68&4.648&0.3385&0.6023\\
&(TIP'24) VRT\cite{VRT}&15&29.64&0.8138&0.3567&0.1197&39.19&5.558&0.2406&0.5015\\
&(CVPR'24) MIA-VSR\cite{MIA-VSR}&15&\makecell[tc]{\cellcolor{red!30}\textbf{31.29}}&\makecell[tc]{\cellcolor{red!30}\textbf{0.8626}}&0.2930&0.0943&\makecell[tc]{\cellcolor{pink!50}\underline{32.15}}&4.631&0.3693&0.5519\\
\cline{1-2}
\multirow{5}{*}{\makecell{Generative \\ Methods}}&(ICCV'21) Real-ESRGAN\cite{RealESRGAN}&15&25.23&0.7147&0.3342&0.0944&52.40&\makecell[tc]{\cellcolor{pink!50}\underline{2.631}}&0.4021&0.5953\\
&(IJCV'24) StableSR\cite{StableSR}&15&25.66&0.7325&0.3155&0.0991&57.94&3.056&0.3733&\makecell[tc]{\cellcolor{red!30}\textbf{0.7187}}\\
&(CVPR'24) Upscale-A-Video\cite{Upscale-A-Video}&15&25.34&0.6776&0.3058&0.1111&61.14&3.127&0.2862&0.5355\\
&(ECCV'24) MGLD-VSR\cite{MGLD-VSR}&15&26.40&0.7118&0.2848&0.0732&34.78&2.887&\makecell[tc]{\cellcolor{pink!50}\underline{0.3905}}&0.5417\\
&SDATC (Ours)&15&26.17&0.7137&\makecell[tc]{\cellcolor{pink!50}\underline{0.2774}}&\makecell[tc]{\cellcolor{red!30}\textbf{0.0636}}&\makecell[tc]{\cellcolor{red!30}\textbf{31.09}}&\makecell[tc]{\cellcolor{red!30}\textbf{2.613}}&\makecell[tc]{\cellcolor{red!30}\textbf{0.4024}}&\makecell[tc]{\cellcolor{pink!50}\underline{0.6119}}\\
\hline

\multirow{5}{*}{\makecell{Non-generative \\ Methods}}&(CVPR'22) BasicVSR++\cite{BasicVSR++}&25&26.85&0.7173&0.4822&0.1861&94.19&6.019&0.1681&0.3637\\
&(CVPR'22) RealBasicVSR\cite{RealBasicVSR}&25&25.95&0.7066&0.3534&0.1028&70.11&\makecell[tc]{\cellcolor{pink!50}\underline{2.839}}&0.3267&0.5252\\
&(ECCV'22) FTVSR\cite{FTVSR}&25&\makecell[tc]{\cellcolor{pink!50}\underline{28.39}}&\makecell[tc]{\cellcolor{pink!50}\underline{0.7802}}&0.4186&0.1649&84.59&5.697&0.2952&0.4611 \\
&(TIP'24) VRT\cite{VRT}&25&26.92&0.7196&0.4833&0.1902&93.75&6.496&0.1706&0.3766\\
&(CVPR'24) MIA-VSR\cite{MIA-VSR}&25&\makecell[tc]{\cellcolor{red!30}\textbf{28.75}}&\makecell[tc]{\cellcolor{red!30}\textbf{0.7871}}&0.4034&0.1638&78.16&5.598&0.3395&0.5314\\
\cline{1-2}
\multirow{5}{*}{\makecell{Generative \\ Methods}}&(ICCV'21) Real-ESRGAN\cite{RealESRGAN}&25&24.89&0.6898&0.3926&0.1191&73.99&2.871&0.3565&0.5385\\
&(IJCV'24) StableSR\cite{StableSR}&25&25.23&0.7022&0.3859&0.1292&74.60&3.440&0.3212&0.5370\\
&(CVPR'24) Upscale-A-Video\cite{Upscale-A-Video}&25&24.70&0.6484&0.3801&0.1213&69.80&3.029&0.2911&\makecell[tc]{\cellcolor{pink!50}\underline{0.5442}}\\
&(ECCV'24) MGLD-VSR\cite{MGLD-VSR}&25&25.47&0.6685&\makecell[tc]{\cellcolor{red!30}\textbf{0.3366}}&\makecell[tc]{\cellcolor{pink!50}\underline{0.0975}}&\makecell[tc]{\cellcolor{pink!50}\underline{55.37}}&2.964&\makecell[tc]{\cellcolor{pink!50}\underline{0.3703}}&0.5001\\
&SDATC (Ours)&25&25.28&0.6705&\makecell[tc]{\cellcolor{pink!50}\underline{0.3488}}&\makecell[tc]{\cellcolor{red!30}\textbf{0.0894}}&\makecell[tc]{\cellcolor{red!30}\textbf{51.00}}&\makecell[tc]{\cellcolor{red!30}\textbf{2.796}}&\makecell[tc]{\cellcolor{red!30}\textbf{0.3796}}&\makecell[tc]{\cellcolor{red!30}\textbf{0.5616}}\\
\hline

\multirow{6}{*}{\makecell{Non-generative \\ Methods}}&(CVPR'22) BasicVSR++\cite{BasicVSR++}&35&24.24&0.6265&0.5852&0.2676&183.37&7.225&0.1037&0.2255\\
&(CVPR'22) RealBasicVSR\cite{RealBasicVSR}&35&23.45&0.6078&0.4722&0.1720&137.54&3.158&0.2715&0.4637\\
&(ECCV'22) FTVSR\cite{FTVSR}&35&\makecell[tc]{\cellcolor{pink!50}\underline{25.13}}&\makecell[tc]{\cellcolor{red!30}\textbf{0.6697}}&0.5436&0.2526&180.28&7.190&0.1983&0.1867\\
&(TIP'24) VRT\cite{VRT}&35&24.26&0.6270&0.5855&0.2686&183.05&7.345&0.1049&0.2379\\
&(CVPR'24) MIA-VSR\cite{MIA-VSR}&35&\makecell[tc]{\cellcolor{red!30}\textbf{25.19}}&\makecell[tc]{\cellcolor{pink!50}\underline{0.6695}}&0.5355&0.2571&223.02&7.325&0.2494&0.3283\\
\cline{1-2}
\multirow{5}{*}{\makecell{Generative \\ Methods}}&(ICCV'21) Real-ESRGAN\cite{RealESRGAN}&35&23.60&0.6200&0.5288&0.2241&172.29&3.993&0.2329&0.4210\\
&(IJCV'24) StableSR\cite{StableSR}&35&23.60&0.6260&0.5367&0.2311&166.80&4.734&0.1576&0.2321\\
&(CVPR'24) Upscale-A-Video\cite{Upscale-A-Video}&35&23.22&0.5893&0.5002&0.1704&115.44&3.229&0.2350&\makecell[tc]{\cellcolor{pink!50}\underline{0.4890}}\\
&(ECCV'24) MGLD-VSR\cite{MGLD-VSR}&35&23.36&0.5719&\makecell[tc]{\cellcolor{pink!50}\underline{0.4272}}&\makecell[tc]{\cellcolor{red!30}\textbf{0.1587}}&\makecell[tc]{\cellcolor{red!30}\textbf{97.90}}&\makecell[tc]{\cellcolor{pink!50}\underline{2.873}}&\makecell[tc]{\cellcolor{pink!50}\underline{0.3275}}&0.4598\\
&SDATC (Ours)&35&22.90&0.5470&\makecell[tc]{\cellcolor{red!30}\textbf{0.4175}}&\makecell[tc]{\cellcolor{pink!50}\underline{0.1602}}&\makecell[tc]{\cellcolor{pink!50}\underline{113.19}}&\makecell[tc]{\cellcolor{red!30}\textbf{2.725}}&\makecell[tc]{\cellcolor{red!30}\textbf{0.3276}}&\makecell[tc]{\cellcolor{red!30}\textbf{0.5545}}\\
\hline

\multicolumn{3}{c|}{Dataset}&\multicolumn{8}{c}{Vid4}\\
\hline
\multirow{5}{*}{\makecell{Non-generative \\ Methods}}&(CVPR'22) BasicVSR++\cite{BasicVSR++}&15&25.74&0.7381&0.3745&0.1566&69.73&5.137&0.2155&0.4110\\
&(CVPR'22) RealBasicVSR\cite{RealBasicVSR}&15&23.92&0.6615&\makecell[tc]{\cellcolor{red!30}\textbf{0.3526}}&\makecell[tc]{\cellcolor{pink!50}\underline{0.1252}}&72.95&\makecell[tc]{\cellcolor{pink!50}\underline{2.933}}&0.2922&0.6185\\
&(TIP'24) VRT\cite{VRT}&15&25.72&0.7418&0.3747&0.1680&70.73&5.939&0.2253&0.4202\\
&(CVPR'24) MIA-VSR\cite{MIA-VSR}&15&\makecell[tc]{\cellcolor{pink!50}\underline{26.40}}&0.7837&0.3739&0.1383&\makecell[tc]{\cellcolor{pink!50}\underline{67.38}}&5.143&0.3321&\makecell[tc]{\cellcolor{pink!50}\underline{0.6883}}\\
&(ECCV'22) FTVSR\cite{FTVSR}&15&26.35&\makecell[tc]{\cellcolor{pink!50}\underline{0.7849}}&0.3634&0.1439&68.93&5.584&0.2947&0.5597\\
&(CVPR'23) CAVSR\cite{FTVSR}&15&\makecell[tc]{\cellcolor{red!30}\textbf{27.33}}&\makecell[tc]{\cellcolor{red!30}\textbf{0.8300}}&0.3665&0.1301&68.95&5.422&0.3039&0.6025\\
\cline{1-2}
\multirow{5}{*}{\makecell{Generative \\ Methods}}&(ICCV'21) Real-ESRGAN\cite{RealESRGAN}&15&22.42&0.6037&0.3838&0.1516&86.16&\makecell[tc]{\cellcolor{red!30}\textbf{2.593}}&0.3336&0.5924\\
&(IJCV'24) StableSR\cite{StableSR}&15&22.15&0.5805&0.3762&0.1430&80.16&3.207&0.3380&0.6460\\
&(CVPR'24) Upscale-A-Video\cite{Upscale-A-Video}&15&21.93&0.5343&0.4134&0.1422&80.23&3.277&\makecell[tc]{\cellcolor{pink!50}\underline{0.3590}}&0.6757\\
&(ECCV'24) MGLD-VSR\cite{MGLD-VSR}&15&22.27&0.5654&0.3741&0.1321&89.46&3.247&0.3529&0.6292\\
&SDATC (Ours)&15&22.49&0.5862&\makecell[tc]{\cellcolor{pink!50}\underline{0.3631}}&\makecell[tc]{\cellcolor{red!30}\textbf{0.1229}}&\makecell[tc]{\cellcolor{red!30}\textbf{65.97}}&3.055&\makecell[tc]{\cellcolor{red!30}\textbf{0.3714}}&\makecell[tc]{\cellcolor{red!30}\textbf{0.7332}}\\
\hline

\multirow{6}{*}{\makecell{Non-generative \\ Methods}}&(CVPR'22) BasicVSR++\cite{BasicVSR++}&25&23.64&0.6210&0.4738&0.2183&137.96&5.621&0.1594&0.2703\\
&(CVPR'22) RealBasicVSR\cite{RealBasicVSR}&25&22.82&0.5931&0.4163&0.1588&116.50&\makecell[tc]{\cellcolor{pink!50}\underline{2.809}}&0.2712&0.5987\\
&(TIP'24) VRT\cite{VRT}&25&23.79&0.6300&0.4717&0.2266&137.68&6.532&0.1663&0.3271\\
&(CVPR'24) MIA-VSR\cite{MIA-VSR}&25&\makecell[tc]{\cellcolor{pink!50}\underline{24.75}}&0.6943&0.4258&0.2019&128.70&5.733&0.2927&0.6024\\
&(ECCV'22) FTVSR\cite{FTVSR}&25&24.70&\makecell[tc]{\cellcolor{pink!50}\underline{0.6980}}&0.4217&0.1984&131.18&6.106&0.2548&0.4861\\
&(CVPR'23) CAVSR\cite{FTVSR}&25&\makecell[tc]{\cellcolor{red!30}\textbf{25.60}}&\makecell[tc]{\cellcolor{red!30}\textbf{0.7389}}&\makecell[tc]{\cellcolor{pink!50}\underline{0.4067}}&0.1849&103.88&5.930&0.2631&0.4682\\
\cline{1-2}
\multirow{5}{*}{\makecell{Generative \\ Methods}}&(ICCV'21) Real-ESRGAN\cite{RealESRGAN}&25&21.96&0.5703&0.4206&0.1672&115.83&\makecell[tc]{\cellcolor{red!30}\textbf{2.662}}&0.3175&0.5899\\
&(IJCV'24) StableSR\cite{StableSR}&25&21.85&0.5561&0.4094&0.1588&93.54&3.416&0.3056&0.6264\\
&(CVPR'24) Upscale-A-Video\cite{Upscale-A-Video}&25&21.49&0.4996&0.4438&0.1566&\makecell[tc]{\cellcolor{red!30}\textbf{86.52}}&3.352&0.3127&\makecell[tc]{\cellcolor{pink!50}\underline{0.6646}}\\
&(ECCV'24) MGLD-VSR\cite{MGLD-VSR}&25&21.77&0.5290&0.4073&\makecell[tc]{\cellcolor{pink!50}\underline{0.1507}}&97.76&3.276&\makecell[tc]{\cellcolor{pink!50}\underline{0.3447}}&0.6051\\
&SDATC (Ours)&25&21.91&0.5362&\makecell[tc]{\cellcolor{red!30}\textbf{0.4055}}&\makecell[tc]{\cellcolor{red!30}\textbf{0.1436}}&\makecell[tc]{\cellcolor{pink!50}\underline{92.56}}&3.157&\makecell[tc]{\cellcolor{red!30}\textbf{0.3666}}&\makecell[tc]{\cellcolor{red!30}\textbf{0.6979}}\\
\hline

\multirow{6}{*}{\makecell{Non-generative \\ Methods}}&(CVPR'22) BasicVSR++\cite{BasicVSR++}&35&21.57&0.4914&0.5838&0.2885&254.62&6.618&0.1114&0.1421\\
&(CVPR'22) RealBasicVSR\cite{RealBasicVSR}&35&20.98&0.4783&0.5229&0.2229&250.18&3.213&0.2326&0.3449\\
&(TIP'24) VRT\cite{VRT}&35&21.62&0.4949&0.5844&0.2907&252.83&7.157&0.1228&0.1806\\
&(CVPR'24) MIA-VSR\cite{MIA-VSR}&35&22.05&0.5357&0.5507&0.2839&348.79&6.824&0.2161&0.3344\\
&(ECCV'22) FTVSR\cite{FTVSR}&35&\makecell[tc]{\cellcolor{pink!50}\underline{22.08}}&\makecell[tc]{\cellcolor{pink!50}\underline{0.5412}}&0.5497&0.2786&302.37&6.898&0.1813&0.1840\\
&(CVPR'23) CAVSR\cite{FTVSR}&35&\makecell[tc]{\cellcolor{red!30}\textbf{22.83}}&\makecell[tc]{\cellcolor{red!30}\textbf{0.5734}}&0.5261&0.2732&298.69&6.986&0.1737&0.2874\\
\cline{1-2}
\multirow{5}{*}{\makecell{Generative \\ Methods}}&(ICCV'21) Real-ESRGAN\cite{RealESRGAN}&35&20.83&0.4874&0.5204&0.2304&235.89&3.213&0.2382&0.4272\\
&(IJCV'24) StableSR\cite{StableSR}&35&20.89&0.4815&0.5186&0.2368&222.96&4.246&0.2102&0.3748\\
&(CVPR'24) Upscale-A-Video\cite{Upscale-A-Video}&35&20.14&0.4095&0.5086&0.2085&\makecell[tc]{\cellcolor{red!30}\textbf{138.97}}&\makecell[tc]{\cellcolor{pink!50}\underline{3.114}}&\makecell[tc]{\cellcolor{pink!50}\underline{0.3240}}&\makecell[tc]{\cellcolor{pink!50}\underline{0.5441}}\\
&(ECCV'24) MGLD-VSR\cite{MGLD-VSR}&35&20.46&0.4392&\makecell[tc]{\cellcolor{pink!50}\underline{0.5023}}&\makecell[tc]{\cellcolor{pink!50}\underline{0.2083}}&\makecell[tc]{\cellcolor{pink!50}\underline{166.07}}&\makecell[tc]{\cellcolor{red!30}\textbf{3.054}}&0.3234&0.4396\\
&SDATC (Ours)&35&20.27&0.4077&\makecell[tc]{\cellcolor{red!30}\textbf{0.4773}}&\makecell[tc]{\cellcolor{red!30}\textbf{0.1919}}&231.08&3.256&\makecell[tc]{\cellcolor{red!30}\textbf{0.3501}}&\makecell[tc]{\cellcolor{red!30}\textbf{0.5994}}\\
\hline
\end{tabular}}
\vspace{-8pt}
\label{tab.1}
\end{table*}

\begin{table*}[!t]
\caption{Quantitative comparison of $\times$4 VSR on different compression level UDM 10 datasets.}
\centering
\footnotesize
\vspace{-2pt}
\renewcommand\arraystretch{0.95}{
\setlength{\tabcolsep}{4.5pt}
\begin{tabular}{c|l|c|c|c|c|c|c|c|c|c}
\hline
\multicolumn{3}{c|}{Dataset}&\multicolumn{8}{c}{UDM10}\\
\hline
\multicolumn{2}{c|}{Metrics}&CRF&PSNR$\uparrow$&SSIM$\uparrow$&LPIPS$\downarrow$&DISTS$\downarrow$&FID$\downarrow$&NIQE$\downarrow$&MANIQA$\uparrow$&CLIP-IQA$\uparrow$\\
\hline

\multirow{5}{*}{\makecell{Non-generative \\ Methods}}
&(CVPR'22) BasicVSR++\cite{BasicVSR++}&15&32.96&0.8936&0.2945&0.1028&39.63&5.914&0.2264&0.4539\\
&(CVPR'22) RealBasicVSR\cite{RealBasicVSR}&15&30.64&0.8762&0.2852&0.1011&51.49&3.852&0.3400&0.4957\\
&(TIP'24) VRT\cite{VRT}&15&33.46&0.9006&0.2850&0.1055&39.15&6.487&0.2335&0.4635\\
&(CVPR'24) MIA-VSR\cite{MIA-VSR}&15&\makecell[tc]{\cellcolor{red!30}\textbf{35.76}}&\makecell[tc]{\cellcolor{red!30}\textbf{0.9384}}&0.2878&\makecell[tc]{\cellcolor{red!30}\textbf{0.0809}}&40.19&5.912&0.3466&0.5863\\
&(ECCV'22) FTVSR\cite{FTVSR}&15&\makecell[tc]{\cellcolor{pink!50}\underline{35.43}}&\makecell[tc]{\cellcolor{pink!50}\underline{0.9374}}&0.2900&0.1005&\makecell[tc]{\cellcolor{red!30}\textbf{37.24}}&6.070&0.3258&0.5463\\
\cline{1-2}
\multirow{5}{*}{\makecell{Generative \\ Methods}}
&(ICCV'21) Real-ESRGAN\cite{RealESRGAN}&15&29.22&0.8691&0.2872&0.1023&52.67&4.354&0.3513&0.5577\\
&(IJCV'24) StableSR\cite{StableSR}&15&28.22&0.8569&\makecell[tc]{\cellcolor{red!30}\textbf{0.2756}}&0.0975&51.62&4.361&0.3808&\makecell[tc]{\cellcolor{pink!50}\underline{0.6538}}\\
&(CVPR'24) Upscale-A-Video\cite{Upscale-A-Video}&15&30.07&0.8498&0.3357&0.1108&58.41&4.631&0.2568&0.4641\\
&(ECCV'24) MGLD-VSR\cite{MGLD-VSR}&15&29.67&0.8515&0.2939&0.1044&46.90&\makecell[tc]{\cellcolor{pink!50}\underline{3.810}}&\makecell[tc]{\cellcolor{pink!50}\underline{0.3887}}&0.5242\\
&SDATC (Ours)&15&29.98&0.8538&\makecell[tc]{\cellcolor{pink!50}\underline{0.2804}}&\makecell[tc]{\cellcolor{pink!50}\underline{0.0940}}&\makecell[tc]{\cellcolor{pink!50}\underline{38.52}}&\makecell[tc]{\cellcolor{red!30}\textbf{3.508}}&\makecell[tc]{\cellcolor{red!30}\textbf{0.3935}}&\makecell[tc]{\cellcolor{red!30}\textbf{0.6606}}\\
\hline

\multirow{5}{*}{\makecell{Non-generative \\ Methods}}&(CVPR'22) BasicVSR++\cite{BasicVSR++}&25&30.93&0.8619&0.3564&0.1403&83.33&6.412&0.1947&0.3315\\
&(CVPR'22) RealBasicVSR\cite{RealBasicVSR}&25&29.00&0.8403&0.3483&0.1272&80.02&3.860&0.3065&0.4445\\
&(TIP'24) VRT\cite{VRT}&25&31.24&0.8679&0.3513&0.1436&82.30&6.928&0.2016&0.3519\\
&(CVPR'24) MIA-VSR\cite{MIA-VSR}&25&\makecell[tc]{\cellcolor{red!30}\textbf{32.55}}&\makecell[tc]{\cellcolor{red!30}\textbf{0.8984}}&\makecell[tc]{\cellcolor{pink!50}\underline{0.3059}}&0.1341&78.14&6.558&0.3292&0.4664\\
&(ECCV'22) FTVSR\cite{FTVSR}&25&\makecell[tc]{\cellcolor{pink!50}\underline{32.27}}&\makecell[tc]{\cellcolor{pink!50}\underline{0.8964}}&0.3306&0.1374&88.71&6.685&0.3015&0.4013\\
\cline{1-2}
\multirow{5}{*}{\makecell{Generative \\ Methods}}&(ICCV'21) Real-ESRGAN\cite{RealESRGAN}&25&28.64&0.8514&0.3323&0.1205&76.86&4.590&0.3139&0.4806\\
&(IJCV'24) StableSR\cite{StableSR}&25&28.01&0.8438&0.3467&\makecell[tc]{\cellcolor{pink!50}\underline{0.1155}}&69.68&4.591&0.3498&\makecell[tc]{\cellcolor{pink!50}\underline{0.5960}}\\
&(CVPR'24) Upscale-A-Video\cite{Upscale-A-Video}&25&28.83&0.8191&0.3816&0.1367&78.42&4.124&0.2588&0.4621\\
&(ECCV'24) MGLD-VSR\cite{MGLD-VSR}&25&28.80&0.8288&0.3330&0.1191&\makecell[tc]{\cellcolor{pink!50}\underline{67.00}}&\makecell[tc]{\cellcolor{pink!50}\underline{3.847}}&\makecell[tc]{\cellcolor{pink!50}\underline{0.3628}}&0.5003\\
&SDATC (Ours)&25&28.88&0.8262&\makecell[tc]{\cellcolor{red!30}\textbf{0.3255}}&\makecell[tc]{\cellcolor{red!30}\textbf{0.1153}}&\makecell[tc]{\cellcolor{red!30}\textbf{66.15}}&\makecell[tc]{\cellcolor{red!30}\textbf{3.524}}&\makecell[tc]{\cellcolor{red!30}\textbf{0.3675}}&\makecell[tc]{\cellcolor{red!30}\textbf{0.6021}}\\
\hline

\multirow{5}{*}{\makecell{Non-generative \\ Methods}}&(CVPR'22) BasicVSR++\cite{BasicVSR++}&35&27.90&0.8062&0.4509&0.2173&163.68&7.267&0.1417&0.2055\\
&(CVPR'22) RealBasicVSR\cite{RealBasicVSR}&35&26.52&0.7841&0.4403&0.1884&165.71&4.235&0.2649&0.3436\\
&(TIP'24) VRT\cite{VRT}&35&27.94&0.8085&0.4501&0.2198&163.22&7.656&0.1455&0.2159\\
&(CVPR'24) MIA-VSR\cite{MIA-VSR}&35&\makecell[tc]{\cellcolor{pink!50}\underline{28.71}}&\makecell[tc]{\cellcolor{pink!50}\underline{0.8356}}&0.4381&0.2186&184.25&7.878&0.2794&0.2866\\
&(ECCV'22) FTVSR\cite{FTVSR}&35&\makecell[tc]{\cellcolor{red!30}\textbf{28.75}}&\makecell[tc]{\cellcolor{red!30}\textbf{0.8363}}&0.4394&0.2106&162.23&7.730&0.2319&0.1706\\
\cline{1-2}
\multirow{5}{*}{\makecell{Generative \\ Methods}}&(ICCV'21) Real-ESRGAN\cite{RealESRGAN}&35&27.12&0.8059&0.4348&0.1937&155.43&5.489&0.2375&0.3138\\
&(IJCV'24) StableSR\cite{StableSR}&35&26.74&0.8017&0.4305&0.1868&149.57&5.630&0.2157&0.3378\\
&(CVPR'24) Upscale-A-Video\cite{Upscale-A-Video}&35&26.71&0.7551&0.4805&0.2188&173.94&4.025&0.2638&\makecell[tc]{\cellcolor{pink!50}\underline{0.4386}}\\
&(ECCV'24) MGLD-VSR\cite{MGLD-VSR}&35&26.77&0.7756&\makecell[tc]{\cellcolor{red!30}\textbf{0.4149}}&\makecell[tc]{\cellcolor{pink!50}\underline{0.1771}}&\makecell[tc]{\cellcolor{red!30}\textbf{117.43}}&\makecell[tc]{\cellcolor{pink!50}\underline{3.998}}&\makecell[tc]{\cellcolor{pink!50}\underline{0.2878}}&0.3808\\
&SDATC (Ours)&35&26.52&0.7474&\makecell[tc]{\cellcolor{pink!50}\underline{0.4278}}&\makecell[tc]{\cellcolor{red!30}\textbf{0.1683}}&\makecell[tc]{\cellcolor{pink!50}\underline{146.33}}&\makecell[tc]{\cellcolor{red!30}\textbf{3.526}}&\makecell[tc]{\cellcolor{red!30}\textbf{0.2958}}&\makecell[tc]{\cellcolor{red!30}\textbf{0.4637}}\\
\hline
\end{tabular}}
\vspace{-8pt}
\label{tab.2}
\end{table*}

\begin{figure*}[!t]
\centering
\includegraphics[width=2\columnwidth]{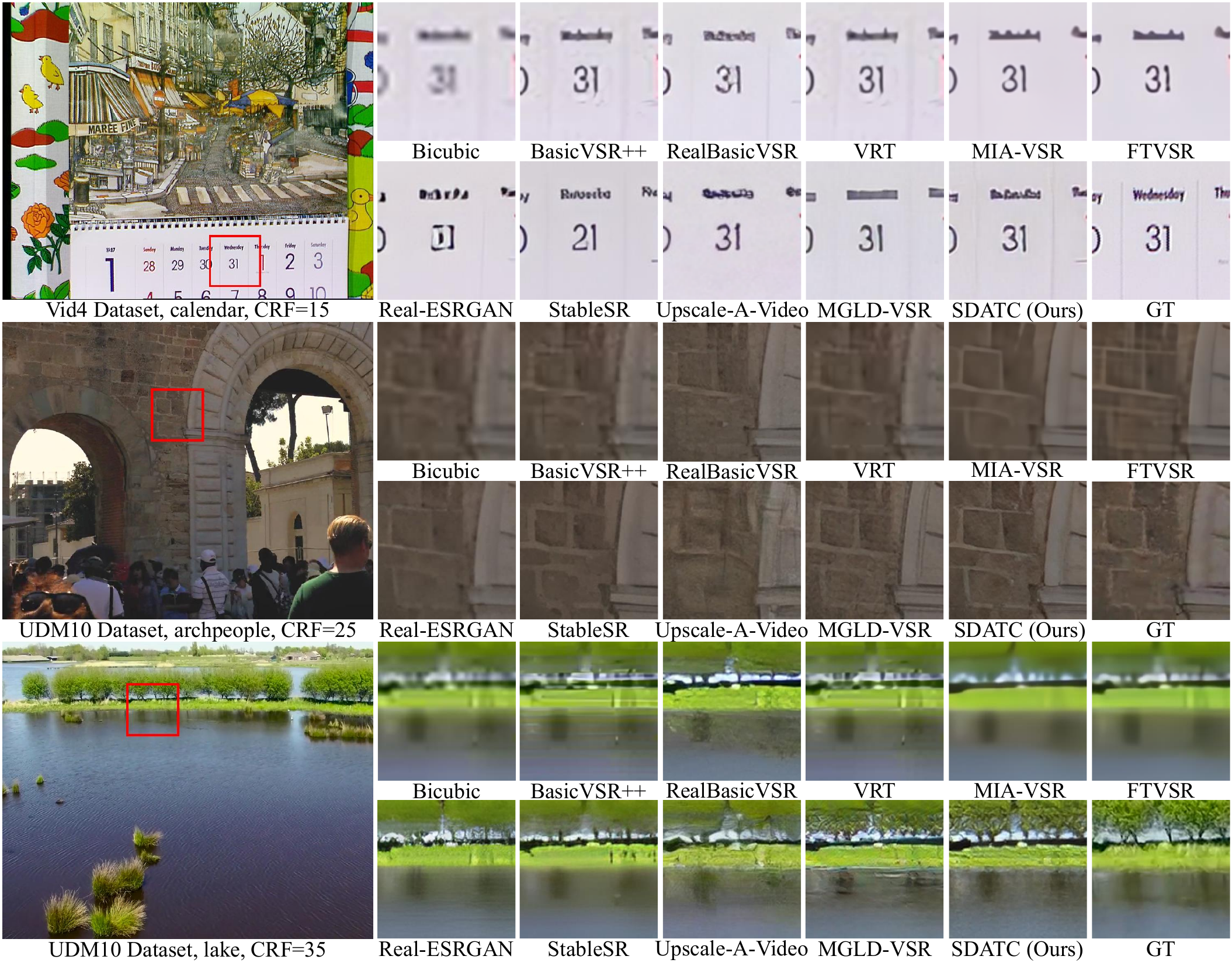}
\vspace{-6pt}
\caption{Qualitative comparison of $\times4$ VSR on different compression level datasets.}
\vspace{-6pt}
\label{fig.6}
\end{figure*}

\begin{figure*}[!t]
\centering
\includegraphics[width=2\columnwidth]{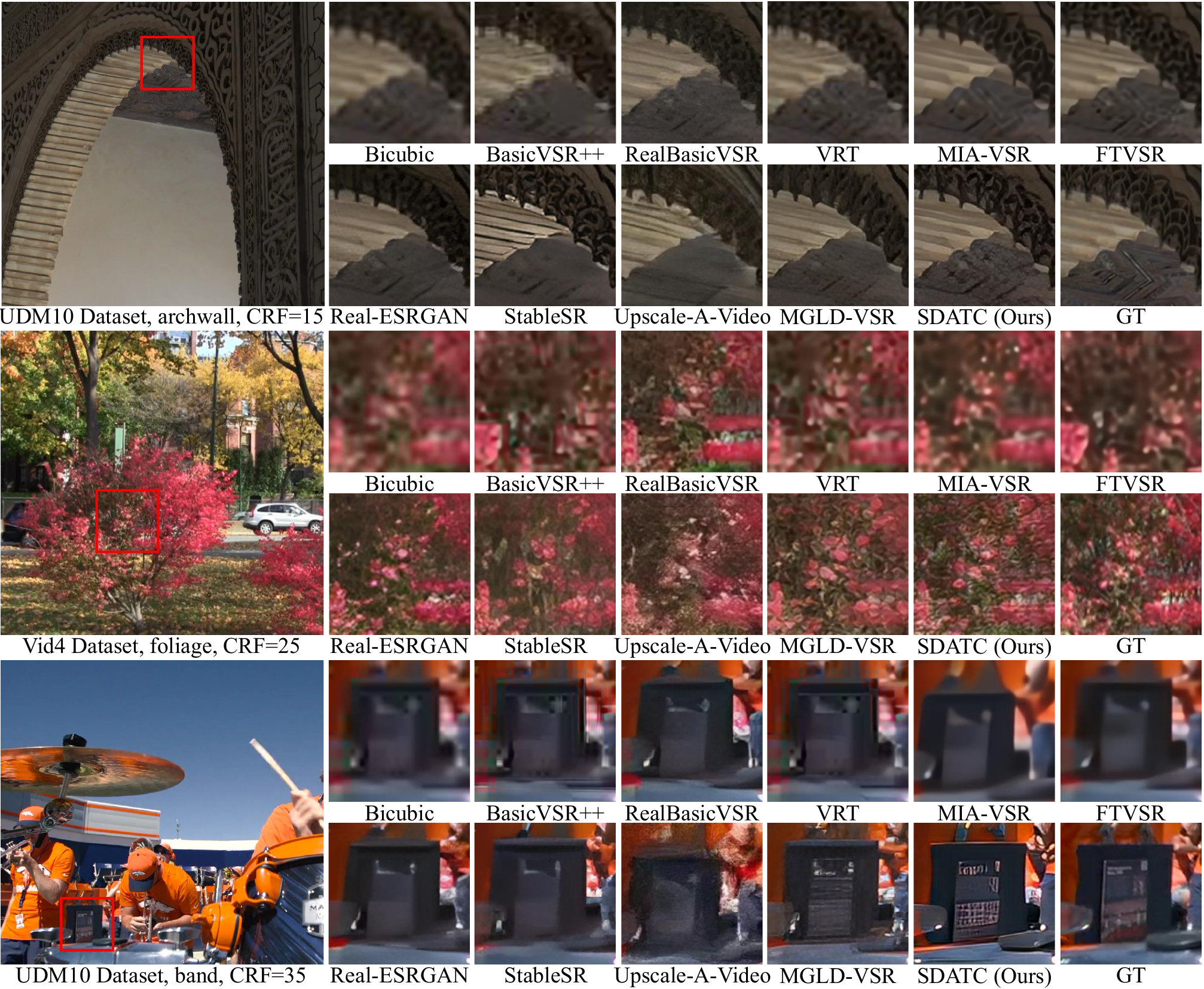}
\vspace{-6pt}
\caption{Qualitative comparison of $\times4$ VSR on different compression level datasets.}
\vspace{-10pt}
\label{fig.7}
\end{figure*}

\section{Experiments}
\subsection{Implementation Details}
\subsubsection{Datasets} 
We train our SDATC on the merged REDS \cite{REDS} training and validation sets, which consist of 266 sequences, each containing 100 frames with a resolution of 1280 $\times$ 720). The remaining 4 sequences (REDS4) are reserved for testing. During training, we utilize the x264 encoder to down-sample and compress videos by a factor of $\times$4. Without loss of generality, we randomly compress videos to bit rates ranging from 10K to 100K. The x264 codec provides different compression levels, i.e., Constant Rate Factor (CRF). The CRF value ranges from 0 (lossless compression) to 51, with 23 being the default value. Following prior works \cite{COMISR,FTVSR,CAVSR}, we select CRFs of 15, 25, and 35 to generate compressed testing videos. Additionally, we evaluate our method on the Vid4 \cite{Vid4} and UDM10 \cite{UDM10} datasets. 

\subsubsection{Training Setting}
The DCM comprises 6 RSTB blocks with a window size of 8. We fine-tune the diffusion model on 8 NVIDIA A100 GPUs. The input clip length is 5, the batch size is 4, and the patch size is 512. The learning rate is initialized as 5 $\times$ 10$^{-5}$ using the Adam \cite{Adam} optimizer. The trade-off parameter $\omega$ is set to 0.75. The noise linear schedule is set to $\eta_1$ = 0.00085 and $\eta_T$ = 0.0120 ($T$ = 1000). During inference, we set 50 sampling steps.

\subsubsection{Evaluation Setting}
We apply various widely utilized reference and non-reference metrics for a comprehensive evaluation, including PSNR, SSIM\cite{SSIM}, LPIPS\cite{LPIPS}, DISTS \cite{DISTS}, FID \cite{FID}, NIQE \cite{NIQE}, MANIQA \cite{MANIQA}, and CLIP-IQA \cite{CLIPIQA}. PSNR and SSIM (Y channel) are reference metrics that measure the similarity between generated images and ground truth images. Other reference metrics, such as LPIPS and DISTS, focus on perceptual quality. FID evaluates the quality of generated images by comparing the feature distributions. For non-reference metrics, NIQE assesses the naturalness of reconstructed images by extracting natural scene statistics features. MANIQA enhances image quality assessment performance through multi-dimension attention mechanisms. CLIP-IQA leverages the vision-language alignment capabilities of the pre-trained CLIP model to evaluate visual quality using text prompts. These metrics effectively measure both image fidelity and perceptual quality, providing an inclusive assessment. Specifically, the emerging non-reference metrics align more closely with human visual perception.

To thoroughly compare the proposed SDATC on the task of compressed VSR, we conduct extensive comparisons with various VSR models (BasicVSR++ \cite{BasicVSR++}, VRT \cite{VRT}, MIA-VSR \cite{MIA-VSR}), compressed VSR models (RealBasicVSR \cite{RealBasicVSR}, Real-ESRGAN \cite{RealESRGAN}, FTVSR \cite{FTVSR}), and diffusion-based models (StableSR \cite{StableSR}, Upscale-A-Video \cite{Upscale-A-Video}, MGLD-VSR \cite{MGLD-VSR}).

\subsection{Quantitative Comparison}
The quantitative experimental results are presented in Tab. \ref{tab.1} and \ref{tab.2}. It is evident that our proposed SDATC comprehensively outperforms other methods in terms of LPIPS, DISTS, FID, NIQE, MANIQA, and CLIP-IQA, at different compression levels on the REDS4, Vid4, and UDM10 datasets. These superior results highlight the effectiveness of the proposed modules and the benefits of incorporating compression-aware generation priors to improve visual perception. Furthermore, SDATC achieves the highest scores in MANIQA and CLIP-IQA, except for the CLIP-IQA value on REDS4 at CRF=15, demonstrating its strong capability in generating realistic details. Similar to other generative approaches, STDAC shows limitations in certain metrics like PSNR and SSIM. This is because these metrics are primarily designed to measure pixel-level fidelity or structural similarity, whereas STDAC focuses on perceptual quality. In other words, diffusion-based methods aim to recover more appealing details but at the expense of fidelity. Notably, STDAC still performs better than other SOTA generative methods in PSNR and SSIM. The comprehensive experimental results deflect SDATC's significant capability to enhance compressed videos and generate realistic details.

\begin{table*}[!htbp]
\caption{Computational Efficiency Comparison. All methods are tested with a 320 $\times$ 180 frame of $\times$4 VSR.}
\centering
\small
\vspace{-2pt}
\renewcommand\arraystretch{0.95}{
\setlength{\tabcolsep}{2.5pt}
\begin{tabular}{c|c|c|c|c|c}
\hline
Non-generative Methods&Trainable Params. / Total Params.&Runtime&Generative Methods&Trainable Params. / Total Params.&Runtime\\
\hline
BasicVSR++\cite{BasicVSR++}&7.0M / 7.0M&2.4s&Real-ESRGAN\cite{RealESRGAN}&16.7M / 16.7M&0.1s\\
RealBasicVSR\cite{RealBasicVSR}&6.3M / 6.3M&0.4s&StableSR\cite{StableSR}&149.9M / 1.5B&50.1s\\
VRT\cite{VRT}&29.1M / 29.1M&0.9s&Upscale-A-Video\cite{Upscale-A-Video}& - / 1.0B &13.3s\\
MIA-VSR\cite{MIA-VSR}& 15.64M / 15.64M & 1.2s &MGLD-VSR\cite{MGLD-VSR}&130.5M / 1.5B&17.5s\\
FTVSR\cite{FTVSR}&10.8M / 10.8M&0.9s&SDATC (Ours)&135.1M / 1.5B&11.6s\\
\hline
\end{tabular}}
\vspace{-10pt}
\label{tab.7}
\end{table*}

\subsection{Qualitative Comparison}
As illustrated in the zoom-in regions of Fig. \ref{fig.6}, the proposed SDATC outperforms CNN-based and Transformer-based methods, such as BasicVSR++, RealBasicVSR, VRT, MIA-VSR, and FTVSR, by producing clearer details. This is particularly evident when the compression degree is high (e.g., CRF$=$35), where other methods yield completely blurry results. Although non-generative methods like FTVSR and MIA-VSR achieve higher PSNR and SSIM values, they tend to produce over-smoothed outcomes. Meanwhile, compared with generative approaches, SDATC restores finer details such as text and numbers in the ``calendar'' sequence and more appealing textures in the ``archpeople'' sequence. It also reconstructs more natural elements like trees, grasslands, and water surfaces in the ``lake'' sequence. Unfortunately, the SOTA diffusion-based VSR method MGLD-VSR introduces grid-like artifacts in the ``archpeople'' sequence and color shift artifacts in the ``lake'' sequence, while Upscale-A-Video produces results with lower fidelity. Additional visual results on different datasets are provided in Fig. \ref{fig.7}, SDATC reconstructs more appealing building structure textures, small texts, and more natural-looking flowers. For non-generative methods, the clarity and level of detail remain insufficient. Although existing generative models can recover objects from compressed frames, the results often appear unrealistic, negatively impacting visual perception. The experimental results across different degradation intensities demonstrate that SDATC excels in both structural rationality and detail clarity. In the presence of severe compression artifacts (e.g., CRF$=$25, 35), SDATC produces finer details and fewer compression artifacts. The effectiveness is attributed to the powerful image understanding and reasoning capabilities of LDMs, as well as the application of LQ image information embedding and structure control.

\subsection{Computational Efficiency Comparison}
The computational efficiency is evaluated and shown in Tab. \ref{tab.7}. Although incorporating pre-trained LDMs into the diffusion-based framework introduces a large number of parameters, we only need to fine-tune limited modules as trainable parameters. The runtime is measured on an NVIDIA A100 GPU. While CNN-based or Transformer-based non-generative methods have fewer training parameters and faster inference times, they struggle to handle VSR tasks with severe compression artifacts. Compared to other generative approaches, our SDATC achieves a lower inference time and remains competitive and computationally affordable.

\begin{figure}[!t]
\centering
\includegraphics[width=0.9\columnwidth]{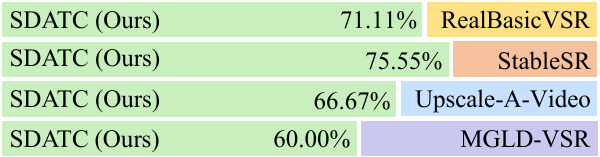}
\vspace{-6pt}
\caption{User study results of ×4 VSR on CRF=25 datasets.}
\vspace{-10pt}
\label{fig.13}
\end{figure}

\subsection{User Study}
We conduct a user study to determine which reconstructed videos were preferred among different methods. Specifically, we invite 15 participants to compare SDATC with RealBasicVSR, StableSR, Upscale-A-Video, and MGLD-VSR in pairwise comparisons. As depicted in Fig. \ref{fig.13}, volunteers prefer the results of SDATC over other approaches on 12 videos.

\begin{table}[!t]
\caption{Ablation study of Distortion Control Module (DCM).}
\centering
\small
\vspace{-2pt}
\renewcommand\arraystretch{0.95}\setlength{\tabcolsep}{4pt}{
\begin{tabular}{c|c|c|c|c}
\hline
Module&DISTS$\downarrow$&NIQE$\downarrow$&MANIQA$\uparrow$&CLIP-IQA$\uparrow$\\
\hline
Baseline&0.1551&4.104&0.1694&0.1470\\
+USM&0.1451&4.107&0.1796&0.1670\\
+DiffBIR&0.1215&3.293&0.2655&0.3169\\
+TMSA&0.1408&3.475&0.2354&0.2486\\
+DCM&\cellcolor{red!30}\textbf{0.1005}&\cellcolor{red!30}\textbf{2.964}&\cellcolor{red!30}\textbf{0.3386}&\cellcolor{red!30}\textbf{0.4371}\\
\hline
\end{tabular}}
\vspace{-6pt}
\label{tab.3}
\end{table}

\begin{figure}[!t]
\centering
\includegraphics[width=\columnwidth]{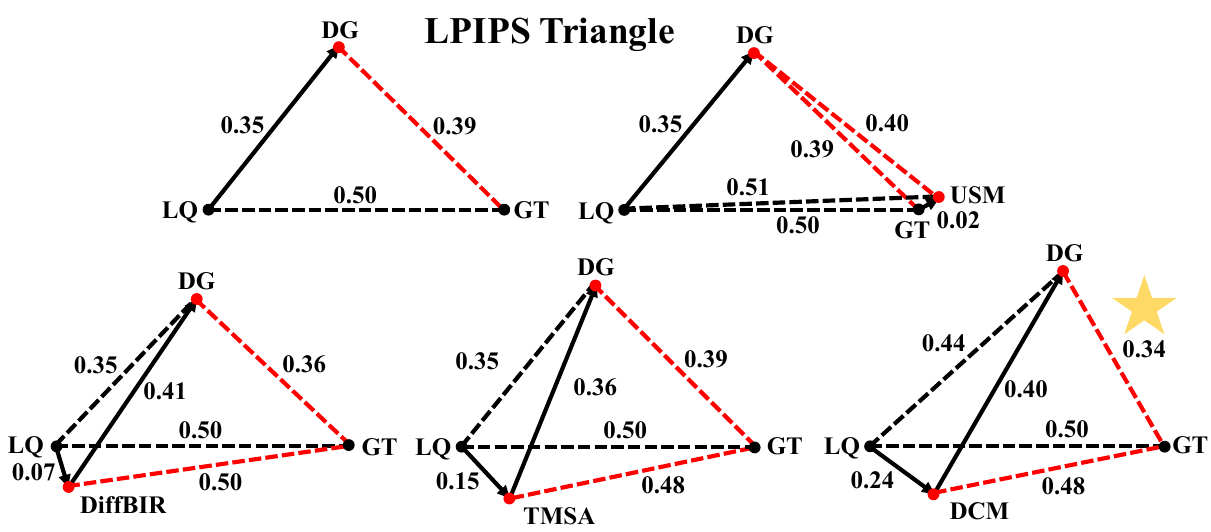}
\vspace{-16pt}
\caption{LPIPS scores of different restoration methods.}
\vspace{-4pt}
\label{fig.8}
\end{figure}

\begin{figure}[!t]
\centering
\includegraphics[width=\columnwidth]{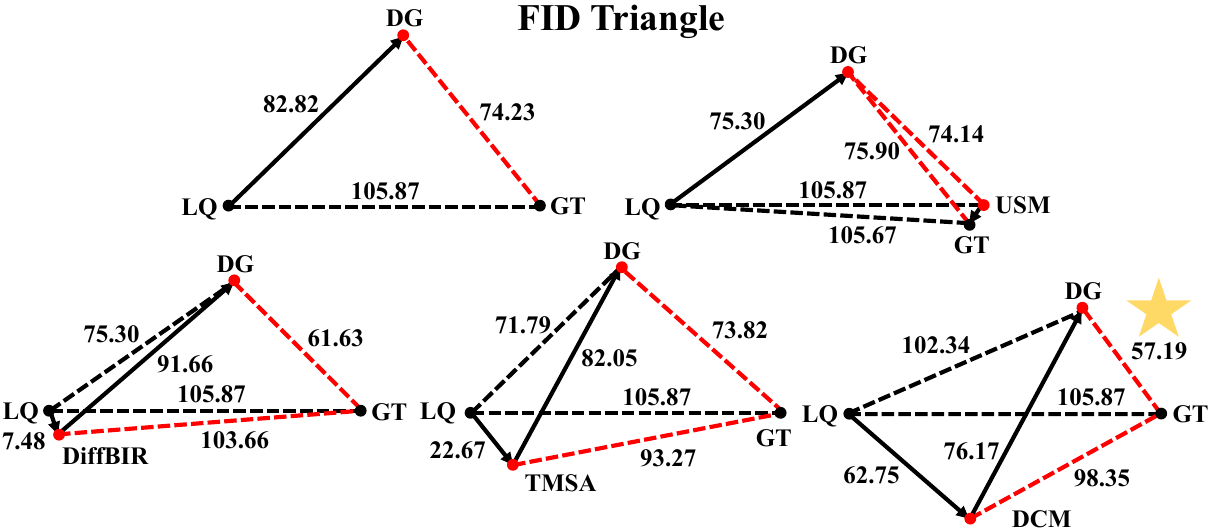}
\vspace{-16pt}
\caption{FID scores of different restoration methods.}
\vspace{-4pt}
\label{fig.9}
\end{figure}

\begin{figure}[!t]
\centering
\includegraphics[width=\columnwidth]{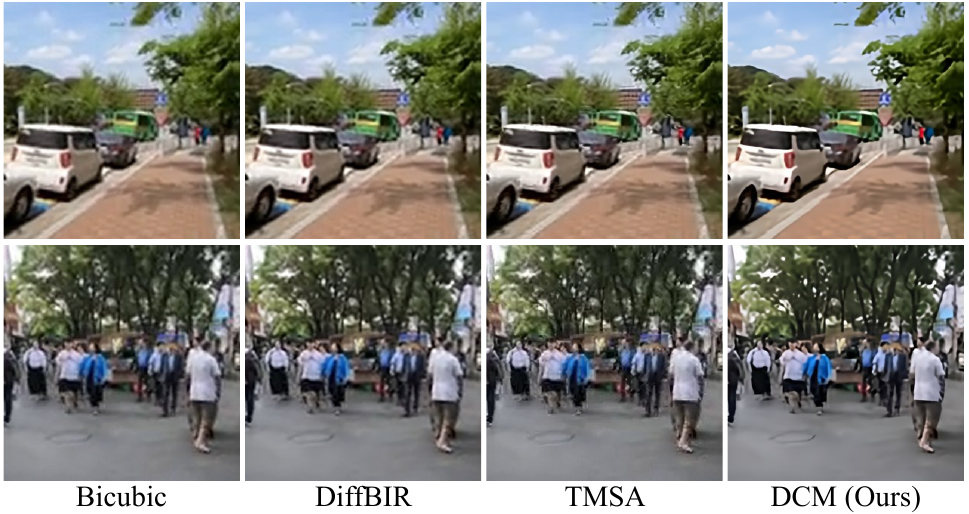}
\vspace{-16pt}
\caption{Visual results of different restoration methods.}
\vspace{-10pt}
\label{fig.10}
\end{figure}

\section{Ablation Study}
In this section, we conduct a detailed analysis of the proposed SDATC diffusion network by evaluating the effectiveness of each module in the spatio-temporal dimension. The experiments are performed on the compressed REDS4 dataset (CRF=25) of $\times$4 VSR. The baseline framework is a default LDMs-based network.

\subsection{Distortion Control Module}
As shown in Tab. \ref{tab.3}, the DCM significantly improves perceptual quality and outperforms other enhancement methods. Specifically, Unsharpen Mask (USM) sharpens GT images to optimize training. DiffBIR \cite{DiffBIR} up-samples images using PixelShuffle and then restores them. TMSA \cite{VRT} extracts multi-frame features before up-sampling. In contrast, DCM dynamically achieves compression discrimination, resulting in higher fidelity generation results. 

To provide an intuitive understanding, we calculate similarity scores for the low-quality (LQ) domain, diffusion generation (DG) domain, GT domain, and enhancement domain of different approaches. The similarity is measured by LPIPS and FID, with lower scores indicating closer distance. As presented in Fig. \ref{fig.8} and \ref{fig.9}, basic diffusion-based VSR up-samples LQ frames by Bicubic and then executes diffusion denoising. It can be observed that DCM achieves the best LPIPS and FID scores between the DG domain and GT domain, meaning that DCM allows the diffusion model to produce outputs most similar to GT frames. Maintaining the generation capacity, DCM enhances spatial fidelity in the diffusion model's results. We also visualize the results of these methods in Fig. \ref{fig.10}, where DCM effectively deduces noises and provides smooth diffusion inputs, benefiting subsequent generation processes.

\begin{table}[!t]
\caption{Ablation study of Compression-Aware Prompt Module (CAPM).}
\centering
\small
\vspace{-2pt}
\renewcommand\arraystretch{0.95}\setlength{\tabcolsep}{2pt}{
\begin{tabular}{c|c|c|c|c}
\hline
Module&PSNR$\uparrow$/SSIM$\uparrow$&NIQE$\downarrow$&MANIQA$\uparrow$&CLIP-IQA$\uparrow$\\
\hline
Baseline&26.26/0.7009&4.104&0.1694&0.1470\\
+Prompt&26.29/0.7006&3.882&0.1867&0.1767\\
+Softmax&26.32/0.7015&3.903&0.1861&0.1820\\
+CAPM&\cellcolor{red!30}\textbf{26.58}/\textbf{0.7110}&\cellcolor{red!30}\textbf{3.803}&\cellcolor{red!30}\textbf{0.2080}&\cellcolor{red!30}\textbf{0.2053}\\
\hline
\end{tabular}}
\vspace{-4pt}
\label{tab.4}
\end{table}

\begin{table}[!t]
\caption{Ablation study of Compression-Aware Prompt Module (CAPM) Artifacts Removal.}
\vspace{-2pt}
\centering
\small
\renewcommand\arraystretch{0.95}{
\begin{tabular}{c|c|c|c}
\hline
\multirow{2}{*}{Module} & \multicolumn{3}{c}{Perception-Sensitive Pixel Loss$\downarrow$} \\
\cline{2-4}
&CRF=15&CRF=25&CRF=35\\ 
\hline
Baseline&0.2348&0.3137&0.5263\\
+CAPM&\cellcolor{red!30}\textbf{0.2058}&\cellcolor{red!30}\textbf{0.2868}&\cellcolor{red!30}\textbf{0.5072}\\
\hline
\end{tabular}}
\vspace{-4pt}
\label{tab.5}
\end{table}

\begin{figure}[!t]
\centering
\includegraphics[width=\columnwidth]{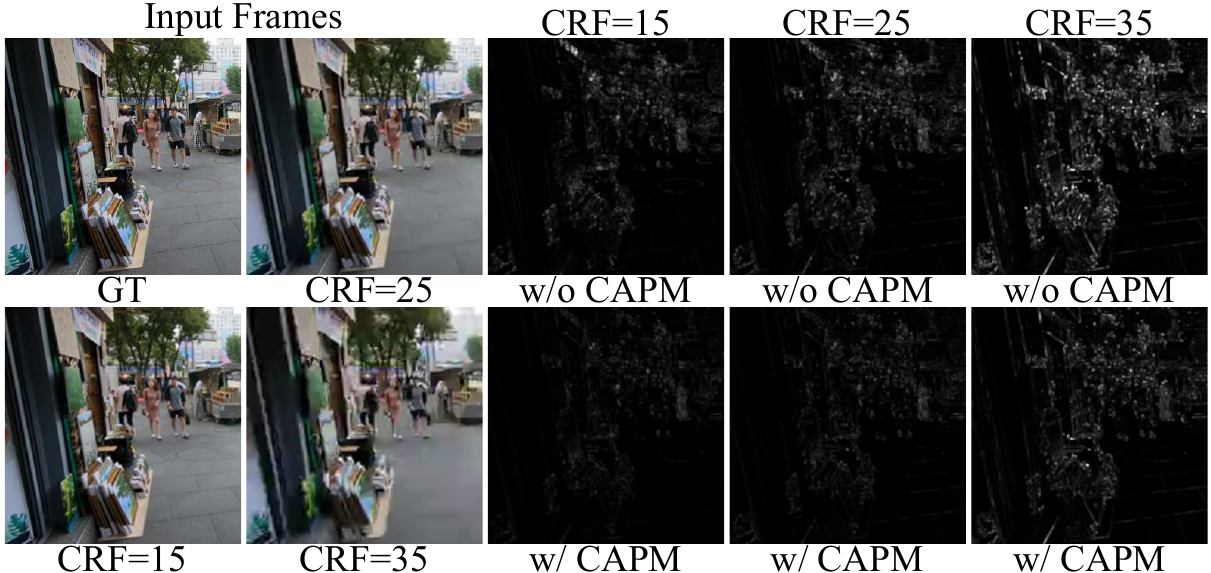}
\vspace{-16pt}
\caption{Visual comparison for the compression artifacts of w/o and w/ CAPM. The bright areas indicate the loss of textural details.}
\label{fig.12}
\vspace{-12pt}
\end{figure}

\subsection{Compression-Aware Prompt Module}
As illustrated in Tab. \ref{tab.4}, the proposed CAPM achieves gains not only in perceptual metrics but also in PSNR and SSIM compared to the baseline. Here, ``+Prompt'' refers to the basic random initialization learnable-prompts, ``+CAPM'' denotes the proposed auxiliary encoding and compression-aware prompts, and ``+Softmax'' indicates a version of CAPM with a Softmax layer during feature extraction. CAPM provides compression-specific prompts to guide reasonable texture generation in both the latent and reconstruction space. Simultaneously, the compression priors extracted from features contribute to improvements in pixel-oriented metrics. Furthermore, we perform an experiment on the REDS4 dataset to quantitatively analyze compressed VSR artifacts. Following the approach of LDL \cite{LDL}, we calculate perception-sensitive pixel loss based on variances. As shown in Tab. \ref{tab.5}, the CAPM module effectively distinguishes between compression and generation artifacts. We also analyzed compression artifacts using perception-sensitive pixel (PSP) loss, which stands for texture details. The PSP loss increases with higher compression levels, but the inclusion of CAPM reduces it.

\begin{table}[!t]
\caption{Ablation study of Spatio-Temporal Attention Module (STAM).}
\centering
\small
\vspace{-2pt}
\renewcommand\arraystretch{0.95}\setlength{\tabcolsep}{3pt}{
\begin{tabular}{c|c|c|c|c}
\hline
Module&VMAF$\uparrow$&NIQE$\downarrow$&MANIQA$\uparrow$&CLIP-IQA$\uparrow$\\
\hline
Baseline&35.70&4.104&0.1694&0.1470\\
+STAM&44.53&3.200&0.3620&0.4869\\
Real-ESRGAN\cite{RealESRGAN}&59.40&2.871&0.3565&0.5385\\
RealBasicVSR\cite{RealBasicVSR}&64.66&2.839&0.3267&0.5252\\
StableSR\cite{StableSR}&58.44&3.440&0.3212&0.5370\\
MGLD-VSR\cite{MGLD-VSR}&56.80&2.964&0.3703&0.5001\\
SDATC (Ours)&\cellcolor{red!30}\textbf{67.22}&\cellcolor{red!30}\textbf{2.796}&\cellcolor{red!30}\textbf{0.3796}&\cellcolor{red!30}\textbf{0.5616}\\
\hline
\end{tabular}}
\vspace{-4pt}
\label{tab.6}
\end{table}

\begin{figure}[!t]
\centering
\includegraphics[width=\columnwidth]{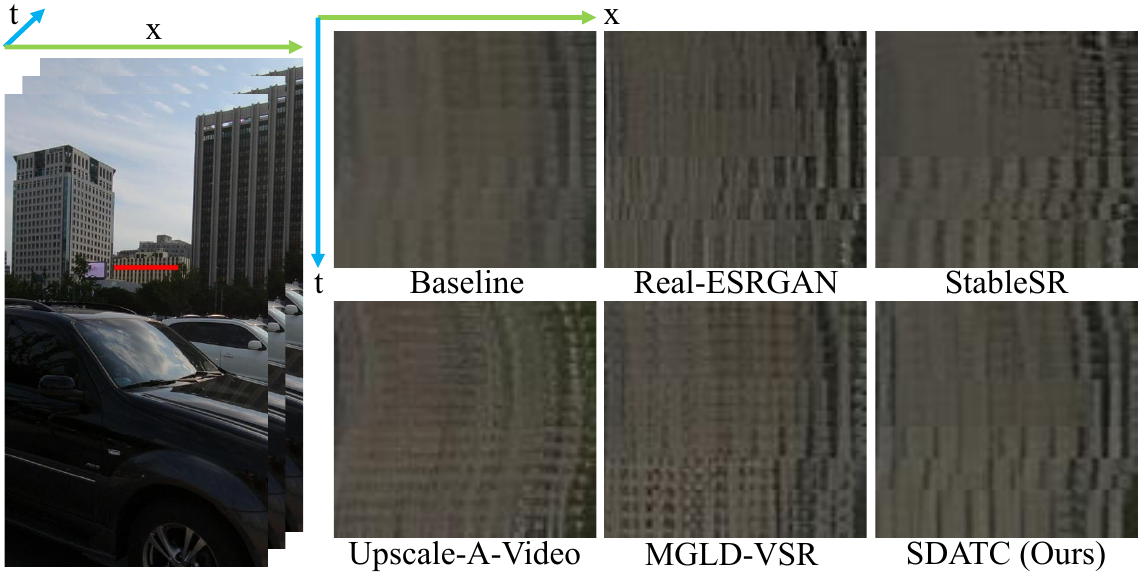}
\vspace{-16pt}
\caption{Temporal profile comparison. The temporal profiles are acquired through concatenating rows at the same location in continuous frames.}
\vspace{-12pt}
\label{fig.11}
\end{figure}

\subsection{Spatio-Temporal Attention Module}
To comprehensively evaluate both spatial quality and temporal coherency, we utilize the video quality assessment metric VMAF \cite{VMAF}, which incorporates motion measures to account for temporal characteristics. From Tab. \ref{tab.6}, the implementation of ``+STAM'' yields superior VMAF scores and perceptual measurements compared to the baseline. Moreover, SDATC overcomes other generation methods in VMAF. The STAM module benefits the spatio-temporal performance. We also demonstrate temporal profiles in Fig. \ref{fig.11} to compare temporal consistency. SDATC achieves smoother multi-frame reconstruction, .

\section{Conclusion}
In this paper, we present a Spatial Degradation-Aware and Temporal Consistent (SDATC) diffusion model for compressed video super-resolution. The key innovation of SDATC lies in leveraging pre-trained diffusion model generation priors and extracting compression priors to enhance reconstruction quality. Specifically, we introduce a distortion control module to modulate diffusion inputs and create controllable guidance, which mitigates the negative impacts of compression in the following denoising process. To further recover compression-lost details, we insert compression-aware prompt modules in latent and reconstruction space to provide adaptive prompts for generation. Finally, we propose a spatio-temporal attention module and optical flow warping to lighten flickering. Extensive experimental evaluations and visual results on benchmark datasets demonstrated the superiority of SDATC over other state-of-the-art methods. Through compression-specific optimizations, we exploited the potential of diffusion models in compressed video super-resolution.

\bibliographystyle{IEEEtran}
\bibliography{references}

\end{document}